%% file: main.tex
\documentclass[11pt,a4paper]{article}
\usepackage{booktabs}
\usepackage[utf8]{inputenc}
\usepackage{geometry}
\usepackage{graphicx}
\usepackage{multirow}
\usepackage{amsmath}
\usepackage{amsthm}
\usepackage{stackengine}
\usepackage{amssymb}
\usepackage{algorithm}
\usepackage{algorithmicx}
\usepackage{algpseudocode} 
\usepackage{makecell}
\usepackage{color}
\usepackage{comment}
\usepackage{mathabx}
\usepackage{blkarray}
\usepackage{mathtools}

\usepackage{caption}
\usepackage{subcaption}

\usepackage[toc]{appendix}

\usepackage[nolist]{acronym}
\begin{acronym}[DRAM]
    \acro{AI}{artificial intelligence}
    \acrodefindefinite{AI}{an}{an}
    \acro{ASIC}{application-specific integrated circuit}
    \acrodefindefinite{ASIC}{an}{an}
    \acro{BD}{bounded delay}
    \acro{BNN}{binarized neural network}
    \acro{CNN}{convolutional neural network}
     \acro{CoTM}{coalesced Tsetlin machine}
    \acro{CTM}{convolutional Tsetlin machine}
    \acro{FSM}{finite state machine}
    \acro{LA}{learning automaton}
    \acrodefplural{LA}{learning automata}
    \acrodefindefinite{LA}{an}{a}
    \acro{ML}{machine learning}
    \acrodefindefinite{ML}{an}{a}
    \acro{TA}{Tsetlin automaton}
    \acro{SSL}{Stochastic Searching on the Line}
    \acrodefplural{TA}{Tsetlin automata}
    \acro{TAT}{Tsetlin automaton team}
    \acro{TM}{Tsetlin machine}
    \acro{RTM}{regression Tsetlin machine}
\end{acronym}

\usepackage{tikz}
\usepackage{pgfplots, pgfplotstable}
\pgfplotsset{compat=1.9}

\usepackage{hyperref}
\hypersetup{
    colorlinks=true,
    linkcolor=blue,
    filecolor=blue,      
    urlcolor=blue,
    citecolor=blue
}

\title{Coalesced Multi-Output Tsetlin Machines with Clause Sharing\footnote{Source code for this paper can be found at \href{https://github.com/cair/PyCoalescedTsetlinMachineCUDA}{https://github.com/cair/PyCoalescedTsetlinMachineCUDA}.}}
\author{Sondre Glimsdal\thanks{Author's status: Lead Researcher at Forsta. E-mail: {\tt sondre.glimsdal@forsta.com}} ~and Ole-Christoffer Granmo\thanks{Author's status: {\it Professor}. The author can be contacted at: Centre for Artificial Intelligence Research (\href{https://cair.uia.no}{https://cair.uia.no}), University of Agder, Grimstad, Norway.  E-mail: {\tt ole.granmo@uia.no}}}

\begin{document}

\maketitle

\input{abstract}

\input{introduction}
\input{tm_basics}
\input{coalesced_tm}
\input{empirical_results}

\input{conclusions}

\bibliographystyle{abbrv}
\bibliography{references}

\input{appendix}

\end{document}

%% file: abstract.tex
\begin{abstract}
Using finite-state machines to learn patterns, \acp{TM} have obtained competitive accuracy and learning speed across several benchmarks, with frugal memory- and energy footprint. A \ac{TM} represents patterns as conjunctive clauses in propositional logic (AND-rules), each clause voting for or against a particular output. While efficient for single-output problems, one needs a separate \ac{TM} per output for multi-output problems. Employing multiple \acp{TM} hinders pattern reuse because each \ac{TM} then operates in a silo. In this paper, we introduce clause sharing, merging multiple \acp{TM} into a single one. Each clause is related to each output by using a weight. A positive weight makes the clause vote for output $1$, while a negative weight makes the clause vote for output $0$. The clauses thus coalesce to produce multiple outputs. The resulting \ac{CoTM} simultaneously learns both the weights and the composition of each clause by employing interacting \ac{SSL} and \acp{TA} teams. Our empirical results on MNIST, Fashion-MNIST, and Kuzushiji-MNIST show that \ac{CoTM} obtains significantly higher accuracy than \ac{TM} on $50$- to $1$K-clause configurations, indicating an ability to repurpose clauses. E.g., accuracy goes from $71.99$\% to $89.66$\% on Fashion-MNIST  when employing $50$ clauses per class (22 Kb memory). While \ac{TM} and \ac{CoTM} accuracy is similar when using more than $1$K clauses per class, \ac{CoTM} 
reaches peak accuracy $3\times$ faster on MNIST with $8$K clauses. We further investigate robustness towards imbalanced training data. Our evaluations on imbalanced versions of IMDb- and CIFAR10 data show that \ac{CoTM} is robust towards high degrees of class imbalance. Being able to share clauses, we believe \ac{CoTM} will enable new \ac{TM} application domains that involve multiple outputs, such as learning language models and auto-encoding.
\end{abstract}

%% file: introduction.tex
\section{Introduction}

\acp{TM} \cite{granmo2018tsetlin} have recently demonstrated competitive accuracy-, memory footprint-, energy-, and learning speed on several benchmarks, spanning tabular data \cite{abeyrathna2021integer,wheeldon2020learning}, images~\cite{granmo2019convtsetlin,sharma2021dropclause}, regression~\cite{abeyrathna2020nonlinear}, natural language~\cite{berge2019text,yadav2021sentiment,yadav2021wordsense,bhattarai2021novelty,yadav2021dwr}, and speech~\cite{lei2021kws}. By not relying on minimizing output error, \acp{TM} are less prone to overfitting. Instead, they use frequent pattern mining and resource allocation principles to extract common patterns in the data. Unlike the intertwined nature of pattern representation in neural networks, a \ac{TM} decomposes problems into self-contained patterns. These are expressed using conjunctive clauses in propositional logic. That is, each pattern is an AND-rule, such as: \textbf{if} input $\mathbf{x}$ \textbf{satisfies} condition $A$ \textbf{and not} condition $B$ \textbf{then} output $y = 1$. The clause outputs, in turn, are combined into a classification decision through a majority vote, akin to logistic regression, however, with binary weights and a unit step output function. Being based on the human-interpretable disjunctive normal form \cite{valiant1984learnable}, like Karnaugh maps \cite{karnaugh1953map}, a \ac{TM} can map an exponential number of input feature value combinations to an appropriate output \cite{granmo2018tsetlin}.

\paragraph{Recent progress on \acp{TM}.} Recent research reports several distinct \ac{TM} properties. The \ac{TM} can be used in convolution \cite{granmo2019convtsetlin}, providing competitive performance on MNIST, Fashion-MNIST, and Kuzushiji-MNIST, in comparison with CNNs, K-Nearest Neighbor, Support Vector Machines, Random Forests \cite{breiman2001random}, XGBoost \cite{chen2016xgboost}, BinaryConnect \cite{courbariaux2015binaryconnect}, Logistic Circuits \cite{LiangAAAI19} and ResNet \cite{he2016deep}. The \ac{TM} has also achieved promising results in text classification \cite{berge2019text,yadav2021dwr}, word sense disambiguation \cite{yadav2021wordsense}, novelty detection \cite{bhattarai2021novelty, bhattarai2021wordlevel}, fake news detection \cite{bhattarai2021fakenews}, semantic relation analysis \cite{saha2020causal}, and aspect-based sentiment analysis \cite{yadav2021sentiment} using the conjunctive clauses to capture textual patterns. Recently, regression \acp{TM} \cite{abeyrathna2020nonlinear} compared favorably with Regression Trees \cite{BreFriOlsSto84a}, Random Forest Regression \cite{breiman2001random}, and Support Vector Regression \cite{NIPS1996_d3890178}.

The above \ac{TM} approaches have further been enhanced by various techniques. By introducing real-valued clause weights, it turns out that the number of clauses can be reduced by up to $50\times$ without loss of accuracy \cite{phoulady2020weighted}. Also, the logical inference structure of \acp{TM} makes it possible to index the clauses on the features that falsify them, increasing inference- and learning speed by up to an order of magnitude \cite{gorji2020indexing}. Multi-granular clauses simplify the hyper-parameter search by eliminating the pattern specificity parameter \cite{gorji2019multigranular}.  In \cite{abeyrathna2021integer}, \ac{SSL} automata \cite{oommen1997stochastic} learn integer clause weights, performing on-par or better than Random Forests \cite{breiman2001random}, XGBoost \cite{chen2016xgboost}, Neural Additive Models \cite{agarwal2020neural}, StructureBoost \cite{lucena2020structureboost}, and Explainable Boosting Machines \cite{nori2019interpretml}.  Closed form formulas for both local and global \ac{TM} interpretation, akin to SHAP \cite{NIPS2017_8a20a862}, were proposed by Blakely et al.~\cite{blakely2021closed}. 

Computationally, \acp{TM} are natively parallel \cite{abeyrathna2021parallel} and  hardware near \cite{wheeldon2020hardware,wheeldon2020learning,wheeldon2021low}, allowing energy usage to be traded off against accuracy by making inference deterministic~\cite{abeyrathna2020deterministic}. Additionally, Shafik et al. show that \acp{TM} can be fault-tolerant, completely masking stuck-at faults~\cite{shafik2020explainability}.

Recent theoretical work proves convergence to the correct operator for ``identity" and ``not". It is further shown that arbitrarily rare patterns can be recognized using a quasi-stationary Markov chain-based analysis. The work finally proves that when two patterns are incompatible, the most accurate pattern is selected \cite{zhang2021convergence}. Convergence for the ``XOR" operator has also recently been proven by Jiao et al.~\cite{jiao2021convergence}.

\begin{figure}[!t]
\centering
\includegraphics[width=5.0in]{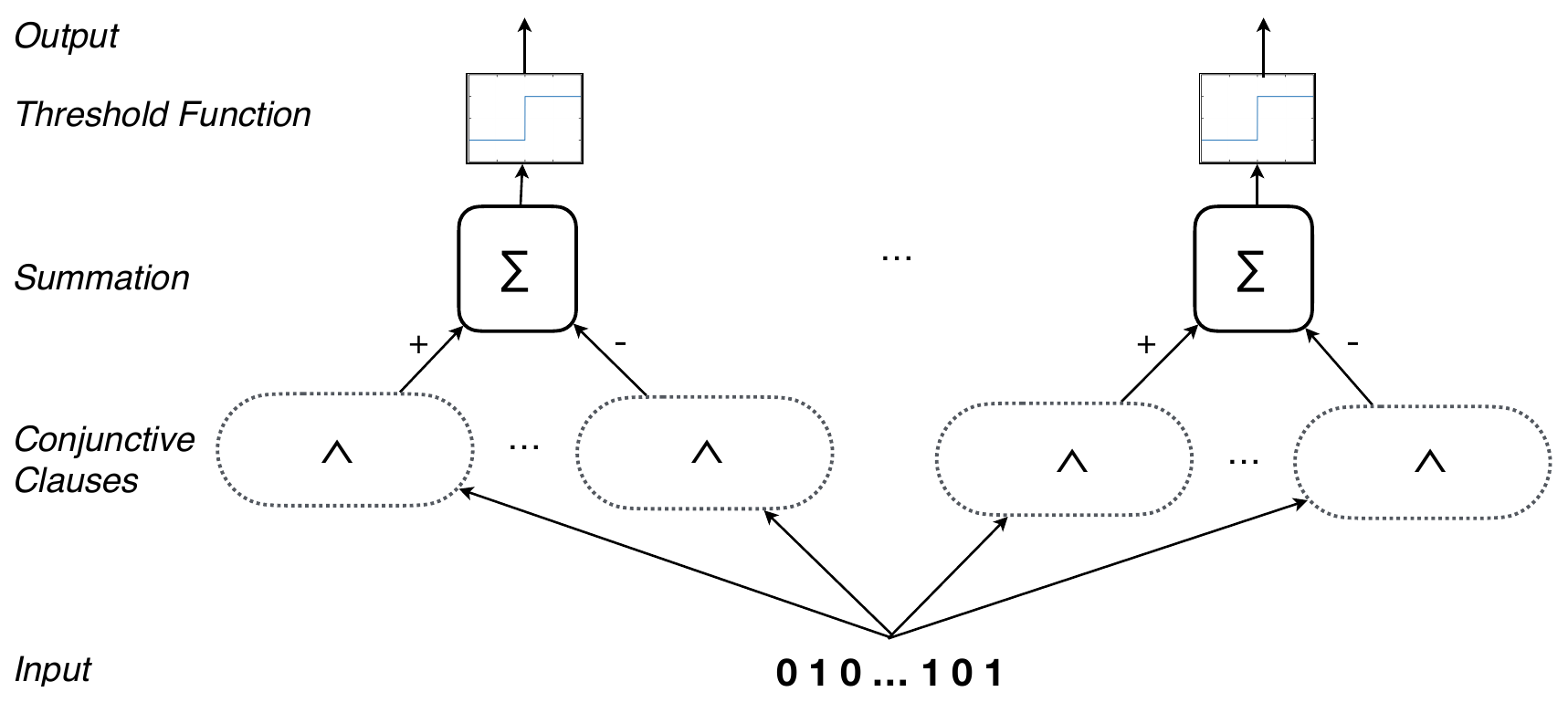}
\caption{The Tsetlin Machine inference structure, introducing clause polarity, a summation operator collecting ``votes", and a threshold function arbitrating the final output \cite{granmo2018tsetlin}.}
\label{figure:architecture_summation}
\end{figure}

\paragraph{Paper Contributions.}
Figure~\ref{figure:architecture_summation} depicts the original multi-output \ac{TM} architecture from~\cite{granmo2018tsetlin}. A \ac{TM} represents patterns as a collection of conjunctive clauses in propositional logic. Each clause vote for or against a particular output, arbitrated by a majority vote. As further depicted in the figure, there is one separate \ac{TM} per output, each maintaining its own set of clauses. This hinders reuse of patterns among the outputs, because each \ac{TM} operates in a silo. 

\begin{figure}[!t]
\centering
\includegraphics[width=4.5in]{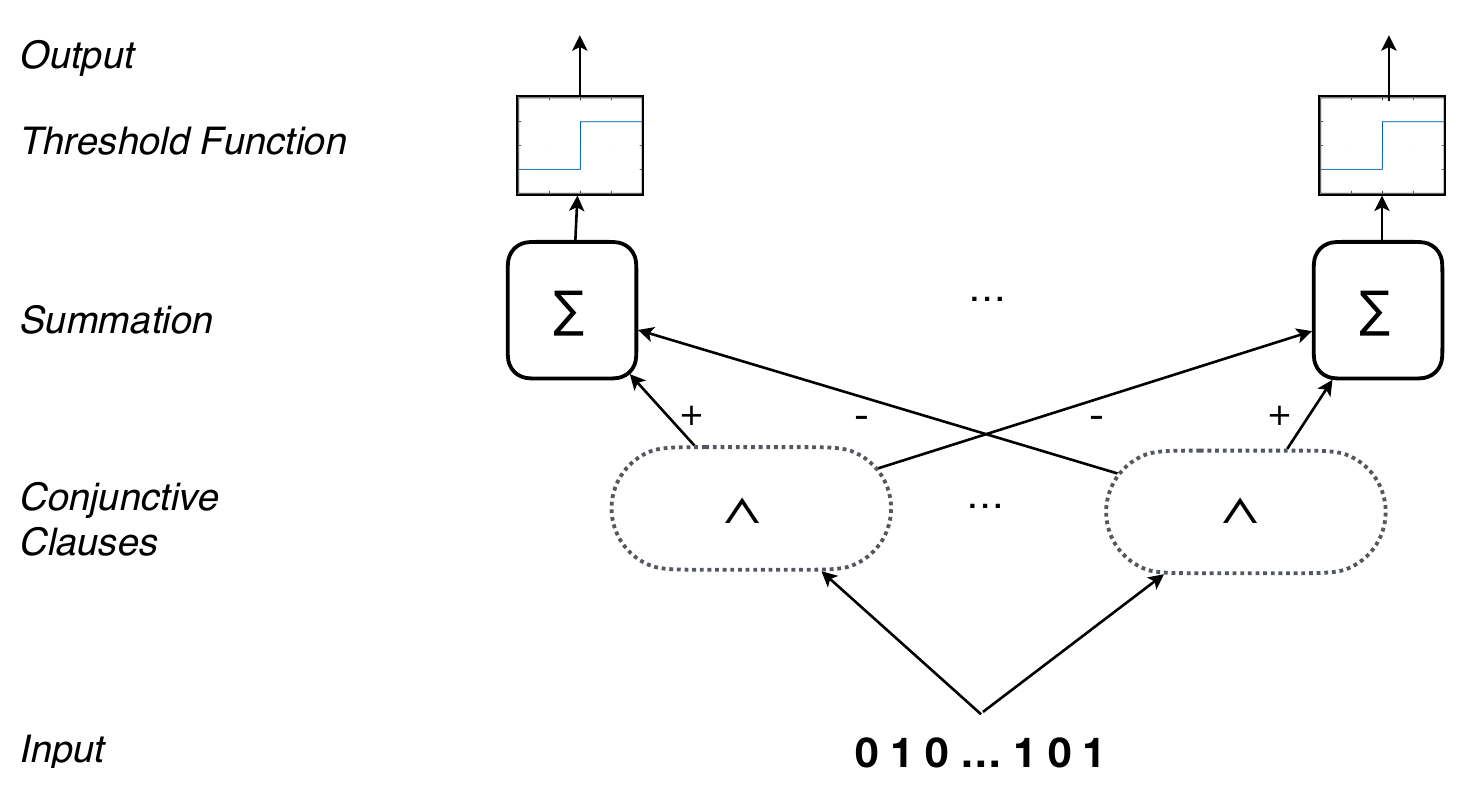}
\caption{The Coalesced Tsetlin Machine (CoTM) inference structure, introducing clause sharing.}
\label{figure:coalesced_architecture_summation}
\end{figure}

In this paper, we introduce a new \ac{TM} architecture that employs a shared pool of clauses, illustrated in Figure~\ref{figure:coalesced_architecture_summation}. Each clause in the pool is related to each output by using a weight. A positive weight makes the clause vote for output $1$, while a negative weight makes it vote for output $0$. The magnitude of the weight decides the impact of the vote. The clauses thus coalesce to produce multiple outputs. The resulting \ac{CoTM} simultaneously learns both the weights and the composition of each clause. The weights are learnt using one \ac{SSL} automaton per clause-output pair. Simultaneously, a team of \acp{TA} configures each clause to maximise output prediction accuracy.

\paragraph{Paper Organization.}

In Section \ref{sec:basics}, we introduce the basics of \ac{TM} inference and learning, providing the basis for \ac{CoTM}. Then, in Section \ref{sec:coalesced_tm}, we describe the \ac{CoTM} in detail, including the architecture for coalescing clauses and learning weights. Our empirical results are presented in Section \ref{sec:empirical_results}, where we evaluate the \ac{CoTM} on 2D Noisy XOR, MNIST, Fashion-MNIST, K-MNIST, CIFAR10, and IMDb comparing with various baselines. We conclude the paper in Section \ref{sec:conclusion} and discuss ideas for further work.

%% file: tm_basics.tex
\section{Tsetlin Machine Basics}\label{sec:basics}
In this section, we provide an introduction to the principles of \ac{TM} inference and learning, which we formalize in Section \ref{sec:coalesced_tm} when presenting the \ac{CoTM}.

\subsection{Inference}

Figure \ref{figure:architecture_summation} depicts a multi-output \ac{TM} architecture. As seen, the architecture uses a simple pattern matching scheme to decide upon the output:

\begin{itemize}
\item One \ac{TM}
is associated with each output, assembling a set of patterns. Each pattern, in turn, is an AND-rule, called a \emph{conjunctive clause}. 

\item The AND-operator ($\land$ in the figure) binds together propositional inputs. These are either False or True ($0$ or $1$ in the figure).

\item For each \ac{TM}, half of the clauses can vote for output value True. We refer to these as \emph{positive} clauses. The other half can vote for output value False. We call these \emph{negative} clauses.

\item  To decide upon a particular output, we count the positive clauses that match the input. Matching negative clauses, on the other hand,  \emph{decrease} the match count.

\item A majority vote decides the final output, indicated by the threshold functions in the figure. If there are more matching negative clauses than positive clauses, the output is False. Otherwise the output is True.
\end{itemize}

As an example, consider the task of determining the sentiment of a collection of movie reviews. To this end, we employ a Set of Words (SoW) representation. That is, each word in the vocabulary is a propositional input, representing the presence or absence of the word in a particular review. Consider, for instance, the review \emph{``The movie was good, and I had popcorn"}. Here, the words ``the", ``movie", ``was", ``good", ``and", ``I", ``had", ``popcorn" are present. These thus take the input value True. Any other words are absent, taking the input value False.

Based on the SoW, the task is to decide upon the nature of each review, whether it is \emph{Positive} or \emph{Negative}. The task thus has two outputs and we consider each output to either be False or True.\footnote{ Note that for this particular task, a movie review cannot be both \emph{Positive} and \emph{Negative} at the same time. Accordingly, we could have modelled the problem as a multi-class problem instead of a multi-output problem~\cite{granmo2018tsetlin}.}

\begin{table}[th]
\center
\begin{tabular}{|c||l||l|}
\hline
\textbf{Clause} & \multicolumn{1}{c||}{\textbf{TM: Positive}} & \multicolumn{1}{c|}{\textbf{TM: Negative}} \\ \hline
\#1            & "good" AND "movie" $\rightarrow$ +1        & "bad" $\rightarrow$  +1                    \\ \hline
\#2            & NOT "bad"  $\rightarrow$ +1                & NOT "good" $\rightarrow$ +1                \\ \hline
\#3            & "horrible" AND "popcorn" $\rightarrow$ -1  & "good" $\rightarrow$ -1                    \\ \hline
\#4            & "bad"  $\rightarrow$ -1                    & "high value" $\rightarrow$ -1              \\ \hline
\end{tabular}
\caption{Eight example clauses for classifying IMDb movie reviews.}\label{table:example_rules}
\end{table}

Table \ref{table:example_rules} contains two example \acp{TM} for the above task, one for output \emph{Positive} and one for output \emph{Negative}. The first column specifies the clause number; the second column contains the clauses for the \ac{TM} assigned to output \emph{Positive}; and the third column covers the \emph{Negative} output \ac{TM}. Clause \#1 and Clause \#2 can vote for output True ('$+1$'-votes), while Clause \#3 and Clause \#4 can vote for output False ('$-1$'-votes). For instance, Clause \#1 for \ac{TM}  \emph{Positive} says that if the input review contains the words ``good" and ``movie", the clause votes for output True. Notice that Clause \#2 consists of a negated input. That is, if the input review \emph{does not} contain the word ``bad", this also counts as a vote towards \emph{Positive} output True. Such negation is useful because one can strengthen the belief in \emph{Positive} from the \emph{absence} of negative sentiment words.

Inference proceeds as follows. Again, consider the example review: \emph{``The movie was good, and I had popcorn"}. To classify this review, we first identify clauses
that are matching the text. For \ac{TM} \emph{Positive}, Clause \#1 matches because both "good" and "movie" are in the text. Clause \#2 also matches the text, since "bad" is absent. Therefore, the score for \ac{TM} \emph{Positive} becomes Clause \#1 + Clause \#2 $= 1 + 1 = 2$. For \ac{TM} \emph{Negative} only Clause \#3 is active, providing a score of $-1$. From a propositional perspective, if a \ac{TM} obtains a non-negative score it outputs True. Otherwise, it outputs False. Accordingly, the output of \ac{TM} \emph{Positive} becomes True, while the output of  \ac{TM} \emph{Negative} becomes False.  

\begin{figure}[ht]
\centering
\includegraphics[width=6.0in]{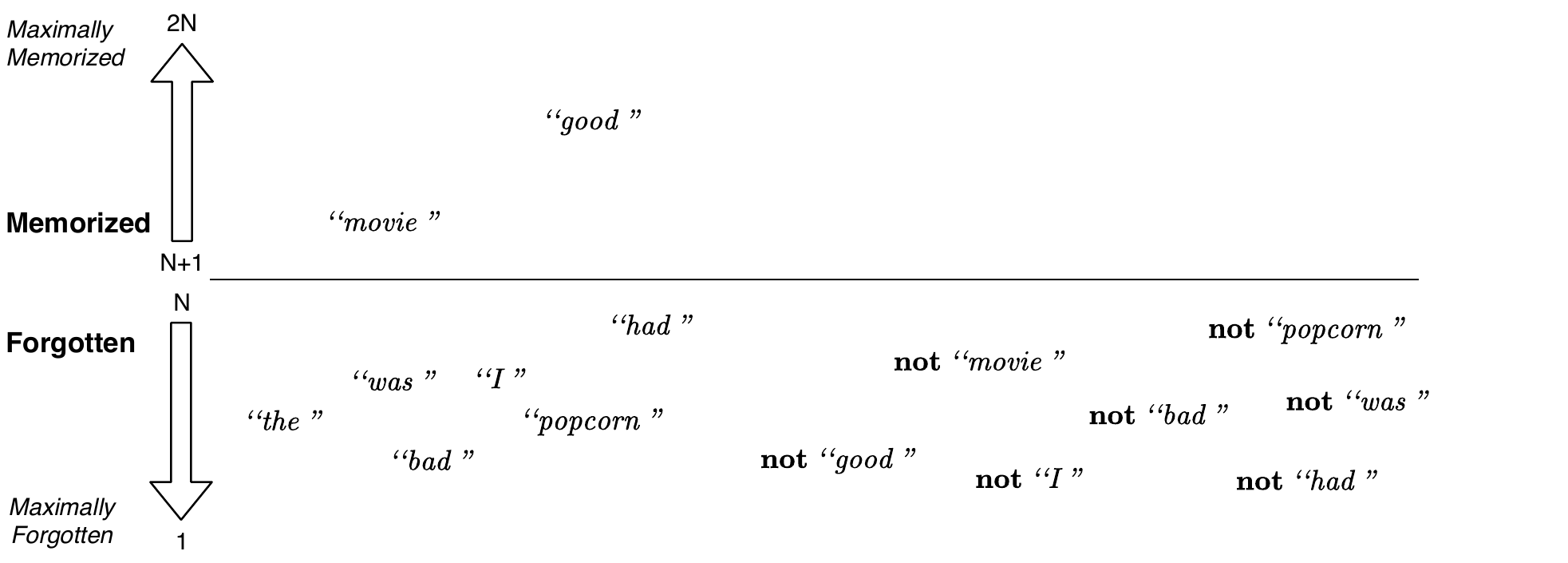}
\caption{Sticky memory with depth $N$ for the clause: ``good" AND ``movie".}
\label{figure:sticky_memory}
\end{figure}

\subsection{Learning}

\paragraph{Sticky Memory.} Memorizing input is the basis of \ac{TM} learning, such as remembering the essence of: \emph{``The movie was good, and I had popcorn"}. To this end, each clause has its own memory where it stores its AND-pattern. In the IMDb case,  the memory contains the truth value of words from the SoW. \ac{TM} \emph{Positive} Clause \#1 stores True for ``movie" and ``good", for instance. However, instead of simply storing the truth values as is, the memory simulates memorization and forgetting. I.e., every time a clause observes a particular truth value, it remembers it longer. Oppositely,  without observations, it eventually forgets the value. Accordingly, each memory entry does not contain a truth value but an integer in the range from $1$ to $2N$. The resulting memory is illustrated in Figure \ref{figure:sticky_memory}. Note that the user sets $N$ to a specific value to control maximal memorization and forgetting. From $N+1$ to $2N$, the word's truth value is part of the clause's AND-pattern. Integer $N+1$ means in memory but easily forgotten. Integer $2N$ means maximally memorized.  A truth value is not part of the pattern in the span $1$ to $N$.  Integer $1$ means maximally forgotten, while integer $N$ means almost memorized. The clause in the figure will, for instance, remember ``good" for a longer time than ``movie" because ``good" is more deeply stored. It is close to memorizing ``had", which is close to $N+1$.\footnote{Note that each memory entry can be seen as a Tsetlin Automaton \cite{Tsetlin1961}, hence the name Tsetlin Machine.}

\begin{figure}[ht]
\centering
\includegraphics[width=6.0in]{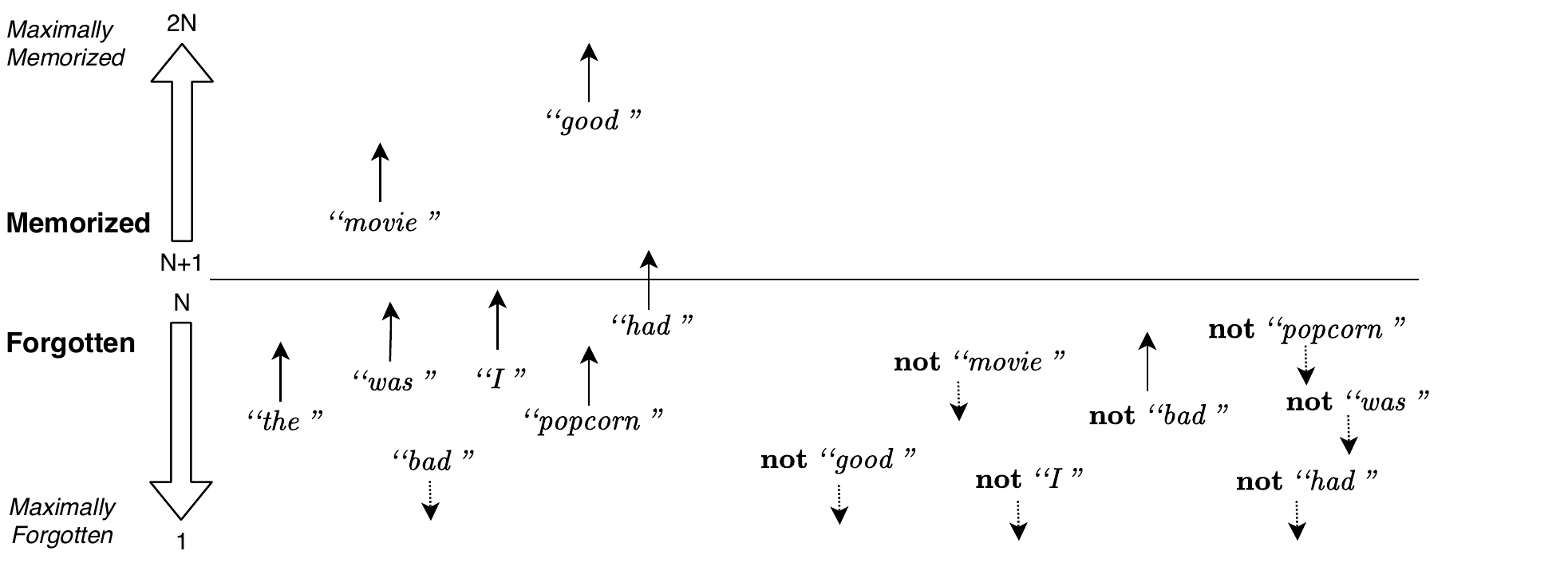}
\caption{The \textbf{memorize}($\mathbf{x}, s$) operator triggered by input \emph{``The movie was good, and I had popcorn."}}
\label{figure:sticky_memory_type_ia}
\end{figure}

\begin{figure}[ht]
\centering
\includegraphics[width=6.0in]{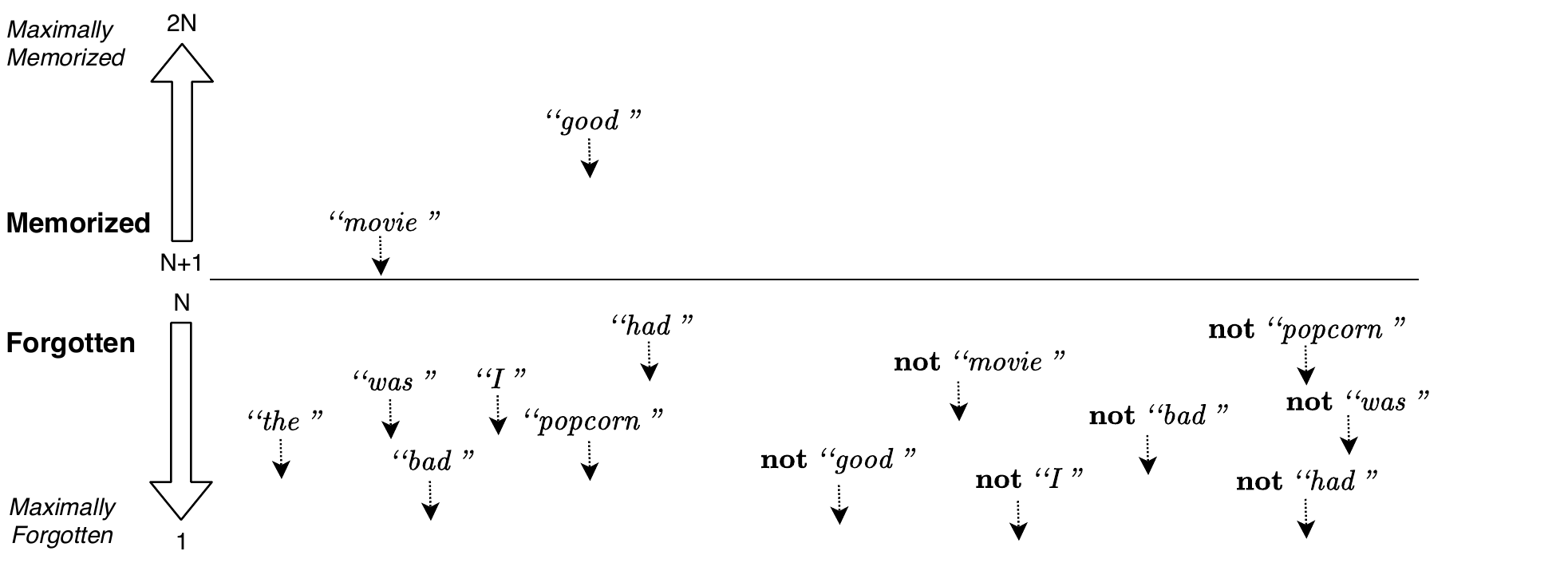}
\caption{The \textbf{forget}($s$) operator triggered by input \emph{``Was good, and I had popcorn."}}
\label{figure:sticky_memory_type_ib}
\end{figure}

\begin{figure}[ht]
\centering
\includegraphics[width=6.0in]{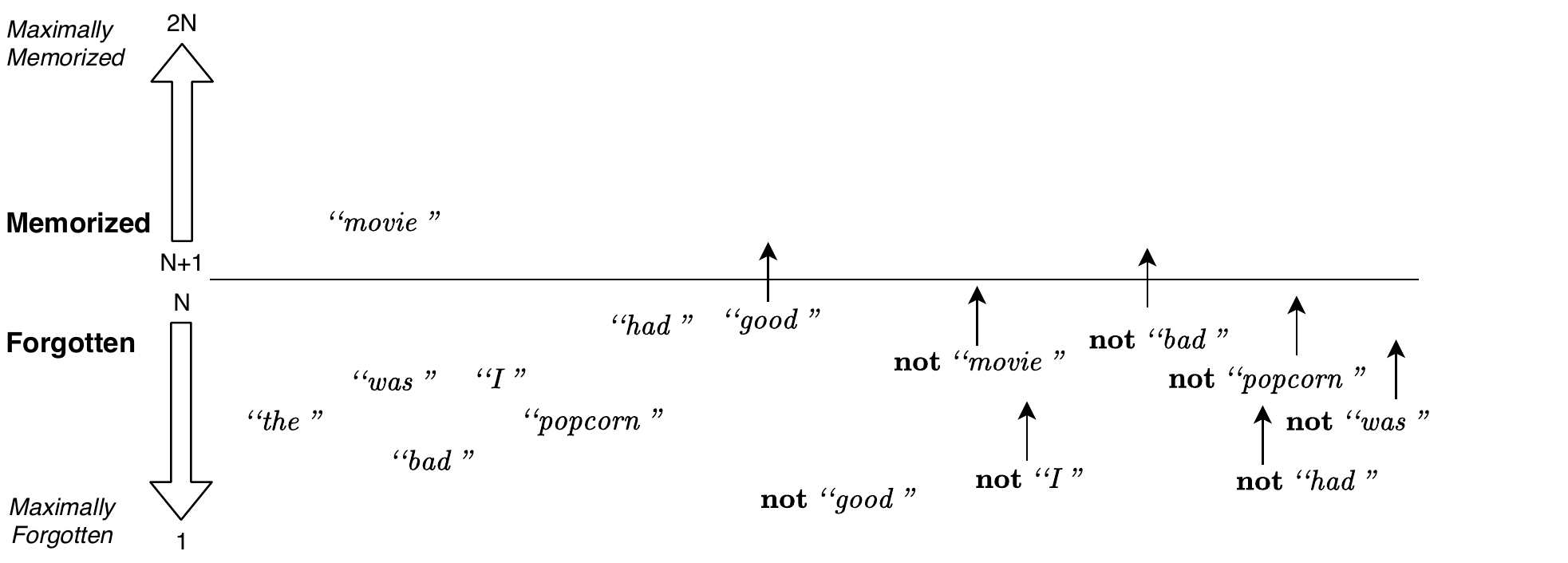}
\caption{The \textbf{invalidate}($\mathbf{x}$) operator applied to input \emph{``The movie was bad, and I had popcorn."}}
\label{figure:sticky_memory_type_ii}
\end{figure}

\paragraph{Memory Updating.} Now, let $\mathbf{x}$ be one particular input to the \ac{TM}. We then have three operators for updating the memory of each clause.
\begin{enumerate}
\item \textbf{memorize($\mathbf{x}, s$)} strengthens the memory of $\mathbf{x}$. However, we do not perform a plain copy. Instead, each truth value in $\mathbf{x}$ increases the  corresponding integer in the memory. As exemplified in Figure \ref{figure:sticky_memory_type_ia}, ``good" in the input increases the integer of ``good". Further, ``bad" is missing in the input, which increases the integer of \textbf{not} ``bad". Simultaneously, the truth values that conflict with the input  have their integers decreased. However, this decrease is randomized, happening with probability $\frac{1}{s}$.  For instance, \textbf{not} ``good" is conflicting with the input. Hence, its integer is decreased randomly with probability $\frac{1}{s}$. The intuition is that memorization must be stronger than forgetting so that patterns can be retained over time. The parameter $s$ is set by the user so that the user can control how quickly truth values are forgotten. In effect, increasing $s$ makes the patterns finer, while decreasing $s$ makes them coarser.

\item \textbf{forget($s$)} performs pure forgetting, as described above. Figure \ref{figure:sticky_memory_type_ib} provides an example. As seen, all the truth values have their integer randomly decreased, with probability $\frac{1}{s}$. Performing this operator several times eventually makes all the truth values maximally forgotten.

\item \textbf{invalidate($\mathbf{x}$)} changes the clause so that it eventually rejects the input $\mathbf{x}$. This operator increases the integer of all False truth values, illustrated in Figure \ref{figure:sticky_memory_type_ii} for input \emph{``The movie was bad, and I had popcorn."} The word ``bad'', for instance, appears in the input. Hence,  \textbf{not} ``bad" gets its integer increased.
\end{enumerate}

\paragraph{Learning Single Clauses.} Four different prediction outcomes guide the learning of each \ac{TM} clause: \begin{itemize}
    \item \textbf{True Positive.} The clause correctly votes for its assigned output. For instance, we have a True Positive outcome when Clause \#1 for \ac{TM} \emph{Positive} matches a review of positive sentiment. In this case, the clause performs memorize($\mathbf{x}, s$). This feedback makes the clause remember and refine the pattern it recognizes in $\mathbf{x}$.
    \item \textbf{False Negative.} The clause fails to vote for its assigned output. If Clause \#1 for \ac{TM} \emph{Positive} does not match a review of positive sentiment, that would be a False Negative outcome. In this case, the clause performs forget($s$). This reinforcement coarsens infrequent patterns, making them frequent.

    \item \textbf{False Positive.} The clause incorrectly votes for its assigned output. If Clause \#1 for \ac{TM} \emph{Positive} matches a review of negative sentiment, we have a False Positive. The clause then performs invalidate($\mathbf{x}$). This feedback makes the clause more discriminative.
    \item \textbf{True Negative}. The clause correctly refrains from voting for its assigned output. That would be the case if Clause \#1 for \ac{TM} \emph{Positive} does not match a review of negative sentiment. This outcome does not trigger any memory updates.
\end{itemize}

\paragraph{Learning Multiple Clauses.} The clauses must learn to coordinate, taking different roles in providing correct output. The \ac{TM} achieves this by introducing a voting margin that we call $t$. We use this parameter to specify how many votes we want the winning output to win by. That is, we want the winning output to win by $t$ votes, but not more. The voting margin makes sure that a sufficient number of clauses support each output. Simultaneously, we do not want the winning output to win by much more than the margin either. That would mean that we use more clauses than necessary. So, the voting margin also ensures prudent usage of the available clauses. Learning of multiple clauses is coordinated as follows:
\begin{enumerate}
\item Obtain next training example. The training example consists of the input truth values $\mathbf{x}$ as well the correct output truth value $y$.
\item Evaluate each clause on input $\mathbf{x}$.
\item Calculate a voting sum for the clauses that evaluate to True (clauses matching~$\mathbf{x}$):
\begin{enumerate}
    \item Add up the votes in favour of $y$, i.e., the correct truth value for the output.
    \item Subtract the votes in favour of $\mathbf{not}~y$, i.e., the incorrect truth value for the output.
    \item We call the summation outcome $v$.
    \item Set $v$ to $t$ if larger than $t$ and to $-t$ if smaller than~$-t$.
\end{enumerate}
\item Go through each clause and update it if $\mathit{Rand}() \le \frac{t-v}{2t}$, where $\mathit{Rand}()$ draws a random value uniformly from the interval $[0,1]$:
\begin{enumerate}
    \item Perform \textbf{memorize}($\mathbf{x}, s$) if the clause matches~$\mathbf{x}$ and belongs to output truth value~$y$. (True Positive)
    \item Perform \textbf{forget}($s$) if the clause \emph{does not} match~$\mathbf{x}$ and belongs to output truth value~$y$. (False Negative)
    \item Perform \textbf{invalidate}($\mathbf{x}$) if the clause matches~$\mathbf{x}$ and belongs to output truth value~$\mathbf{not}~y$. (False Positive)
\end{enumerate}
\end{enumerate}
Observe that if we are far from achieving the voting margin for a particular example, we update clauses more aggressively. E.g., if the voting sum is $-t$ or smaller, we update all the clauses. If the voting sum is zero, we randomly update each clause with probability $0.5$. If we are close to $t$, we calm down the updating. When reaching or surpassing $t$, we update none of the clauses. In this way, clauses individually and gradually assign themselves to training examples that have not yet reached the voting margin. As a result, inference accuracy increases over time as the clauses specialize on different subsets of the training examples.

%% file: coalesced_tm.tex
\begin{figure}[!ht]
\centering
\includegraphics[width=6.0in]{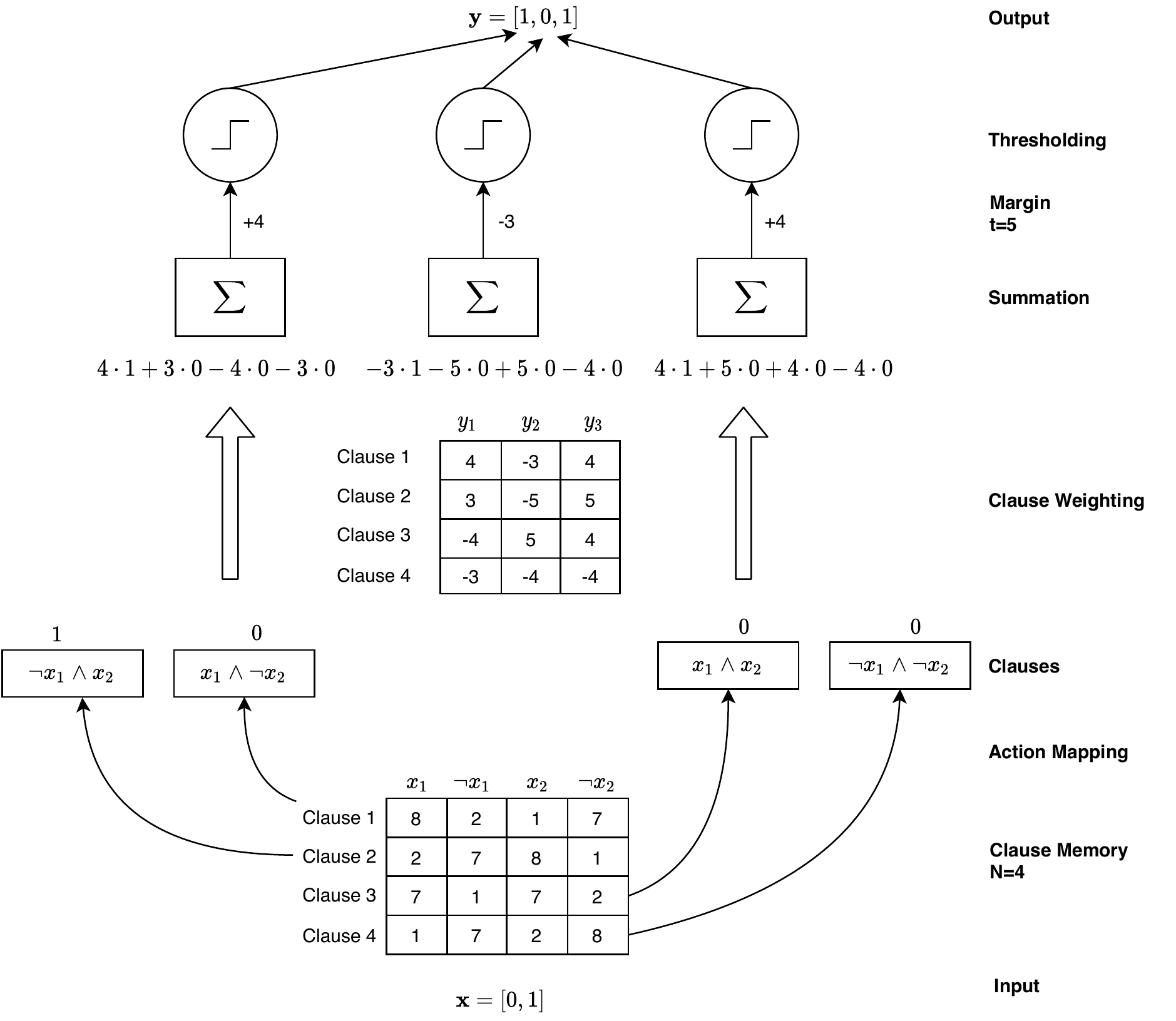}
\caption{The Coalesced Tsetlin Machine structure with inference example.}\label{figure:coalesced_tm}
\end{figure}

\section{Coalesced Tsetlin Machine}\label{sec:coalesced_tm}

We here introduce the \ac{CoTM}, formulated in terms of propositional- and linear algebra. While we cover all the key structures and formulas here, the full details are found in Appendix \ref{appendix:a}.

\subsection{Structure}

A \ac{CoTM} can be defined as a quadruple:
\begin{equation}
\{\mathcal{X}, \mathcal{Y}, \mathcal{C}, \mathcal{W}\}.
\end{equation}
We start with introducing each element of the quadruple, with reference to Figure \ref{figure:coalesced_tm}.
\begin{itemize}
    \item The \textbf{input space} of the \ac{CoTM} is denoted $\mathcal{X}$. The input space consists of vectors $\mathbf{x}$ of $o$ propositional inputs:  $\mathbf{x}=[x_1,\ldots,x_o] \in \mathcal{X}, \mathcal{X} = \{0,1\}^{o}$. In the figure, we have two propositional inputs $\mathbf{x}=[x_1, x_2] \in \mathcal{X}, \mathcal{X} = \{0,1\}^{2}$.
    
    \item The \textbf{output space} is denoted $\mathcal{Y}$. The output space contains vectors $\mathbf{y}$ of $m$ propositional outputs: $\mathbf{y} = [y^1, y^2, \ldots, y^m] \in \mathcal{Y}, \mathcal{Y} = \{0,1\}^{m}$. In the figure, we have three outputs $\mathbf{y}=[y_1, y_2, y_3] \in \mathcal{Y}, \mathcal{Y} = \{0,1\}^{3}$.
    
    \item The \textbf{memory space} $\mathcal{C} = \{1, 2, \ldots, 2N\}^{n \times 2o}$ is the space of memory matrices $C$. A memory matrix $C$ formalizes the sticky memory from Section \ref{sec:basics}. Each of the $n$ rows represents the pattern memory of a single clause. The columns, in turn, represent the $o$ inputs in $\mathbf{x}$ and their negations. Together, the inputs and their negations are referred to as literals. In the figure, we have four literals: $x_1, \lnot x_1, x_2$, and $\lnot x_2$. With four clauses, four literals, and memory depth $N=4$, we get the memory space $\mathcal{C} = \{1, 2, \ldots, 8\}^{4 \times 4}$.

    \item  The \textbf{weight space} $\mathcal{W} = \{\ldots, -2, -1, 1, 2, \ldots \}^{m \times n}$ consists of weight matrices~$W$. A weight matrix $W$ relates each clause in $C$ to an output in $\mathbf{y}$. Consider the output and clause of a specific entry in the weight matrix. A positive weight assigns the clause to output value $1$. A negative weight assigns it to output value $0$. The magnitude of the weight decides the impact of the assignment. With the three outputs and four clauses in the figure, we get the weight space $W \in \mathcal{W} = \{\ldots, -2, -1, 1, 2, \ldots \}^{3 \times 4}$. As seen, Clause~\#1 is assigned output values $y_1=1$, $y_2=0$, and $y_3=1$.
\end{itemize}
Observe how the pattern matrix $C$ combined with the weight matrix $W$ configure a complete \ac{CoTM}.

\subsection{Output Prediction}

Based on the above quadruple, we can predict the output $\mathbf{y}$ from input $\mathbf{x}$ using propositional- and linear algebra as shown in Eqn. \ref{eqn:prediction_1}:
\begin{equation}
    \hat{\mathbf{y}} = U(W \cdot \mathit{And}(\mathit{Imply}(G(C), \mathbf{x})))^T.\label{eqn:prediction_1}
\end{equation}
Above, $\hat{\mathbf{y}}$ is the multi-output prediction. As specified, it is calculated from the input $\mathbf{x}$ using the following operators:
\begin{itemize}
    \item $G(\cdot)$ maps the memory matrix $C$ element-wise to \emph{Exclude} and \emph{Include} actions, respectively $0$ and $1$. Each entry is mapped using the function:
    \begin{equation}
        g(c) = \begin{cases}
        0, \mathbf{if}~ 1 \le c \le N\\
        1, \mathbf{if}~ N+1 \le c \le 2N.
    \end{cases}
    \end{equation}

    \item $\mathit{Imply}$ is an element-wise logical \emph{imply} operator ($\Rightarrow$), which implements the \emph{Exclude} and \emph{Include} actions. It takes a matrix of Exclude/Include actions $[a_{j,k}]$ as input, with $j$ referring to a clause and $k$ referring to a literal.  If $a_{j,k}$ is $0$ (Exclude), then $a_{j,k} \Rightarrow x_k$ is always $1$. Accordingly, the $x_k$ value does not impact the result. Conversely, the value of $a_{j,k} \Rightarrow x_k$ is decided solely by the $x_k$ value if $a_{j_k}$ is~$1$ (Include).

    \item $\mathit{And}$ is a row-wise AND operator. The operator takes a matrix of truth values as input. For each row, it ANDs together all the truth values of that row.

    \item $U$ is an element-wise unit step thresholding operator. $U$ takes a vector as input and applies the unit step function $v \ge 0$ on each vector entry.
\end{itemize}
These operators are fully specified in Appendix \ref{appendix:a}, while Figure \ref{figure:coalesced_tm} provides a prediction example.

In the figure, we trace the input $\mathbf{x}=[0,1]$ through each operator. The clause memory in the figure is of depth $N=4$. So, for example, Clause \#1 includes $x_1$ and $\lnot x_2$ because their respective memory matrix entries are $8$ and $7$ (larger than or equal to $5$). It excludes $\lnot x_1$ and $x_2$, having memory entries $2$ and $1$ (equal to or smaller than~$4$).  Notice how the memory and weight matrices implement XOR for $y_1$, AND for $y_2$, and OR for $y_3$. Hence, input $\mathbf{x} = [0,1]$ provides output $\mathbf{y} = [1,0,1]$.

\subsection{Updating of Memory Matrix}

The memory matrix is updated based on a training example $(\mathbf{x}, \mathbf{y})$ using three kinds of feedback matrices: $F_i^{Ia}$, $F_i^{Ib}$, and $F_i^{II}$. Further, which clauses are eligible for feedback are compiled in the matrices $R^I$ and $R^{II}$. The clauses are randomly selected according to the voting margin of each output values $y_i$ in $\mathbf{y}$ as described in Section~\ref{sec:basics}. In brief, the memory matrix  $C_{T+1}$ for time step $T+1$ is calculated from the memory matrix $C_T$ of time step $T$:
\begin{equation}
    C^*_{T+1} = C_T + \sum_{i=1}^m \left(Q^{II}_i \circ F^{II}_i + Q^I_i \circ F^{Ia}_i - Q^I_i \circ F^{Ib}_i\right).
\end{equation}
As seen, each output $y_i$ of output vector $\mathbf{y}$ is considered one at a time. Both $F_i^{Ia}$ and $F_i^{II}$ increments memory entries, while $F_i^{Ib}$ decrements them. The matrix $Q^I_i$ maps the $i$-rows of matrix $R^I$ from the clause level to the literal level. Similarly, $Q^{II}_i$ maps the $i$-rows of matrix $R^{II}$. These single out which clauses to update using the Hadamard product $\circ$. The final step is to clip the entries in $C$ to make sure that they stay within $1$ and $2N$:
\begin{equation}
    C_{T+1} = \mathit{clip}\left(C^*_{T+1}, 1, 2N\right).
\end{equation}

The feedback matrixes can be summarized as follows:
\begin{itemize}
    \item \textbf{Type Ia Feedback.} This feedback matrix is denoted $F_i^{Ia}$ and concerns output $y_i$ of output vector $\mathbf{y}$. It operates on clauses that both match the input $\mathbf{x}$ and that are assigned to output value $y_i$ (by a negative clause-output weight for $y_i = 0$ and a positive clause-output weight for $y_i=1$). $F_i^{Ia}$ only affects literals that are True and tunes the clauses to represent the current input $\mathbf{x}$ more finely, reinforcing \emph{Include} actions.
    
    \item \textbf{Type Ib Feedback.} This feedback matrix is denoted $F_i^{Ib}$ and again concerns output~$y_i$.  It operates on all the clauses that are assigned to output value $y_i$ by the clause-output weights. For those clauses that \emph{does not} match the input $\mathbf{x}$, it operates on all of the literals. For clauses that match the input, it only affects False literals. In effect, the matrix coarsens the clauses by making them forget literals, reinforcing \emph{Exclude} actions. 
    
    \item \textbf{Type II Feedback.} This feedback matrix is denoted $F_i^{II}$ and concerns output $y_i$. It operates on clauses that both match the input $\mathbf{x}$ and that are assigned to the negated output value, $\mathbf{not}~y_i$. Only literals that are False are affected. This matrix increases the discrimination power of the matching clauses by introducing literals that invalidate the matching.
    
\end{itemize}
Together,  $F_i^{Ia}$ and  $F_i^{Ib}$ implement \textbf{forget} and \textbf{memorize} from Section \ref{sec:basics}. $F_i^{II}$  implements \textbf{invalidate}. See Appendix \ref{appendix:a} for a formal definition of the feedback matrices. 

\begin{figure}[!ht]
\centering
\includegraphics[width=2.5in]{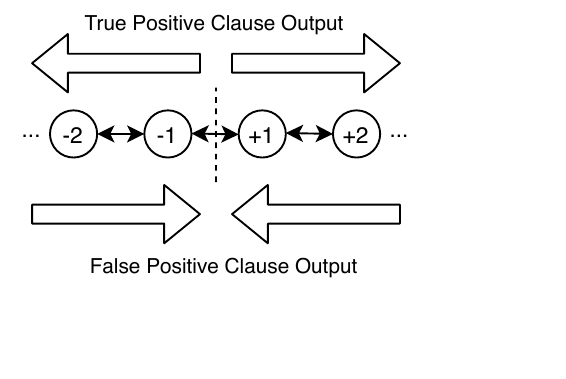}
\caption{Type Ia Feedback changes the weights in one-step increments away from $0$. Type II Feedback changes the weights in one-step decrements towards $0$.}\label{figure:weight_updating}
\end{figure}

\subsection{Updating of Weight Matrix}

As shown in Figure \ref{figure:weight_updating}, the updating of weights is quite straightforward. Weights are only updated in the case of Type Ia or Type II feedback.  Type Ia Feedback changes the weights in one-step increments away from $0$. Type II Feedback changes the weights in one-step decrements towards $0$. Type Ib Feedback leaves the weights unchanged. The following equation captures the updating of the weight matrix from time step $T$ to $T+1$:
\begin{equation}
W_{T+1} = W_T + (R^I + R^{II}) \circ
\left(\begin{bmatrix}
c_1 y_1&c_2 y_1&\cdots&c_n y_1 \\
c_1 y_2&c_2 y_2&\cdots&c_n y_2\\
\vdots&\vdots&\ddots&\vdots\\
c_1 y_m&c_2 y_m&\cdots&c_n y_m\\
\end{bmatrix}-
\begin{bmatrix}
c_1 \bar y_1&c_2 \bar y_1&\cdots&c_n \bar y_1\\
c_1 \bar y_2&c_2 \bar y_2&\cdots&c_n \bar y_2\\
\vdots&\vdots&\ddots&\vdots\\
c_1 \bar y_m& c_2 \bar y_m &\cdots&c_n \bar y_m
\end{bmatrix}\right).
\end{equation}
The matrices $R^I$ and $R^{II}$ are non-overlapping, singling out which clause rows $j$ are eligible for Type I and Type II Feedback. For this purpose, we again use the Hadamard product $\circ$. Let $c_j, j \in \{1, \ldots, n\}$ refer to the output of the clause in row $j$ in $C$. Together with the output value $y_i$ the clause output value controls the direction of the weight update.

%% file: empirical_results.tex
\section{Empirical Results}\label{sec:empirical_results}

In this section, we evaluate the \ac{CoTM} on six different datasets.

\paragraph{2D Noisy XOR.} The 2D Noisy XOR dataset contains $4 \times 4$ binary images, $2500$ training examples and $10~000$ test examples. The image bits have been set randomly, except for the $2 \times 2$ patch in the upper right corner, which reveals the class of the image. A diagonal line is associated with class $1$, while a horizontal or vertical line is associated with class $0$. Thus the dataset models a 2D version of the XOR-relation. Furthermore, the dataset contains a large number of random non-informative features to measure susceptibility towards the curse of dimensionality. To examine robustness towards noise we have further randomly inverted $40\%$ of the outputs in the training data.

\textbf{IMDB.} The IMDb dataset contains $50~000$ highly polar movie reviews for binary sentiment classification~\cite{maas-EtAl:2011:ACL-HLT2011}.

\textbf{CIFAR10.} The CIFAR-10 dataset consists of $60~000$ $32 \times 32$ colour images. There are $10$ classes with $6~000$ images per class \cite{Krizhevsky09learningmultiple}.

\textbf{MNIST.} The MNIST dataset has been used extensively to benchmark machine learning algorithms, consisting of $28\times 28$ grey scale images of handwritten digits~\cite{lecun1998gradient}.

\textbf{Kuzushiji-MNIST.} This dataset contains $28 \times 28$ grayscale images of Kuzushiji characters, cursive Japanese. Kuzushiji-MNIST is more challenging than MNIST because there are multiple distinct ways to write some of the characters  \cite{Clanuwat2018}.

\textbf{Fashion-MNIST.} This dataset contains $28 \times 28$ grayscale images of articles from the Zalando catalogue, such as t-shirts, sandals, and pullovers  \cite{xiao2017}. This dataset is quite challenging, with a human accuracy of $83.50$\%.

The latter three datasets contain $60~000$ training examples and $10~000$ test examples. We binarize these datasets using an adaptive Gaussian thresholding procedure with window size $11$ and threshold value $2$. Accordingly, the \ac{CoTM} operates on images with 1 bit per pixel. All experiments are repeated $10$ times and we report average results for the last $25$ epochs, unless otherwise noted. We ran the \ac{CoTM} on a NDVIDIA Tesla V100 GPU, calculating the clause outputs and updates in parallel.

\begin{table*}[ht]
\centering
\begin{tabular}{c|c|c|c|c|c|c|c|c} 
\#Clauses per class&50&100&250&500&1K&2K&4K&8K\\
Voting Margin&625&1250&3125&6.25K&12.5K&25K&5K&10K\\
Specificity&10.0&10.0&10.0&10.0&10.0&10.0&5.0&5.0\\
\hline
Weighted, Ac. (\%)&97.86&98.43&98.82&98.98&99.14&\textbf{99.22}&\textbf{99.28}&\textbf{99.33}\\
Coalesced, Ac. (\%)&\textbf{98.33}&\textbf{98.80}&\textbf{99.03}&\textbf{99.14}&\textbf{99.18}&99.21&99.26&99.29\\
\hline
Training Time (s)&21.6&21.0&21.1&21.3&20.9&24.1&42.6&86.7\\
Testing Time (s)&1.7&1.6&1.6&1.6&1.5&1.6&4.0&6.5\\
Model Size (kB)&22&44&110&220&439&879&1758&3516\\
\end{tabular}
\caption{Weighted Convolutional \ac{TM} and Convolutional \ac{CoTM} MNIST mean test accuracy and execution time per epoch, for an increasing number of clauses.}\label{table:mnist_clauses_vs_accuracy_and_execution_time}
\end{table*}

\paragraph{MNIST Results.}  We consider scalability first, investigating how the number of clauses affects performance.  Table \ref{table:mnist_clauses_vs_accuracy_and_execution_time} reports performance on unaugmented MNIST, for various number of clauses per class. We see that \ac{CoTM} outperforms \ac{TM} accuracy-wise for $50$ to $1000$ clauses. Indeed, \ac{CoTM} operates at a similar level as \ac{TM} with half the number of clauses, indicating an ability to repurpose clauses towards multiple classes. For $2000$  to $8000$ clauses, \ac{CoTM} performs competitively. 
As further seen in the table, going from $50$ clauses to $8000$ clauses (a $\times 160$ increase) increases testing time $3.8$ times. Similarly, training time increases $4$ times. Finally, model size increases proportionally with the number of clauses. Employing $50$ clauses per class gives a model of size $22$ kB, providing a test accuracy of $98.26$. As the number of clauses increases, so does test accuracy, with $8000$ clauses per class giving a test accuracy of $99.29$.

\paragraph{Fashion-MNIST and Kuzushiji-MNIST Results.} We observe similar behaviour for Fashion-MNIST in Table~\ref{table:fmnist_clauses_vs_accuracy_and_execution_time} and  Kuzushiji-MNIST in Table~\ref{table:kmnist_clauses_vs_accuracy_and_execution_time}. However, the accuracy difference between the weighted \ac{TM} and \ac{CoTM} is even larger for these datasets.

\begin{table*}[ht]
\centering
\begin{tabular}{c|c|c|c|c|c|c|c|c} 
\#Clauses per class&50&100&250&500&1K&2K&4K&8K\\
Voting Margin&625&1250&3125&6.25K&12.5K&2.5K&5K&10K\\
Specificity&15.0&15.0&15.0&15.0&15.0&15.0&15.0&15.0\\
\hline
Weighted, Ac. (\%) &82.33&83.73&88.25&88.79&89.42&89.89&90.65&91.18\\
Coalesced, Ac. (\%)&\textbf{86.79}&\textbf{87.4}&\textbf{88.0}&\textbf{89.20}&\textbf{89.83}&\textbf{90.00}&\textbf{90.71}&91.18\\
\hline
Training Time (s)&25.1&24.1&24.3&23.3&23.7&25.3&45.6&81.9\\
Testing Time (s)&1.6&1.6&1.6&1.5&1.7&1.6&4.1&6.4\\
Model Size (kB)&22&44&110&220&439&879&1758&3516\\
\end{tabular}
\caption{Weighted Convolutional \ac{TM} and Convolutional \ac{CoTM} Fashion-MNIST mean test accuracy and execution time per epoch, for an increasing number of clauses.}\label{table:fmnist_clauses_vs_accuracy_and_execution_time}
\end{table*}

\begin{table*}[!!h]
\centering
\begin{tabular}{c|c|c|c|c|c|c|c|c} 
\#Clauses per class&50&100&250&500&1K&2K&4K&8K\\
Voting Margin&625&1250&3125&6.25K&1.25K&2.5K&5K&10K\\
Specificity&10.0&10.0&10.0&10.0&10.0&10.0&10.0&10.0\\
\hline
Weighted, Ac. (\%) &71.99&89.19&92.75&93.86&94.89&95.40&95.85&96.08\\
Coalesced, Ac. (\%)&\textbf{89.66}&\textbf{91.92}&\textbf{93.71}&\textbf{94.65}&\textbf{95.05}&\textbf{95.77}&\textbf{96.17}&\textbf{96.33}\\
\hline
Training Time (s)&24.2&23.4&23.4&23.7&23.7&25.6&47.3&85.1\\
Testing Time (s)&1.7&1.6&1.6&1.6&1.6&1.6&4.1&6.4\\
Model Size (kB)&22&44&110&220&439&879&1758&3516\\
\end{tabular}
\caption{Weighted Convolutional \ac{TM} and Convolutional \ac{CoTM}  Kuzushiji-MNIST mean test accuracy and execution time per epoch, for an increasing number of clauses.}\label{table:kmnist_clauses_vs_accuracy_and_execution_time}
\end{table*}

\begin{figure}[ht]
\centering
\pgfplotstableread{data/mnist_stats_8000_10000_5.00.txt}{\mnist}
\pgfplotstableread{data/mnist_5.0_80000_10000_10.txt}{\cmnist}
\begin{tikzpicture}
\tikzstyle{more densely dashed}=[dash pattern=on 5pt off 1pt]
	\begin{axis}[
	    ymin=95, ymax=100,
	    xmin=0, xmax=100,
		xlabel=Epoch,
		xtick distance=5,
		ytick distance=0.5,
		ylabel=Accuracy (\%),
        width=0.8\textwidth,
        height=0.6\textwidth,
        legend pos=south east, legend cell align={left},
        ymajorgrids=true, grid style={dotted,gray},
        xmajorgrids=true, grid style={dotted,gray}
	]
	\addplot [black, dashed] table {\mnist};
	\addplot [black, densely dotted] table {\cmnist};
	\addlegendentry{Weighted TM}
	\addlegendentry{Coalesced TM}
	\end{axis}
\end{tikzpicture}
\caption{Single-run test accuracy per epoch for \ac{CoTM} and \ac{TM} on MNIST.}
\label{figure:accuracy_per_epoch}
\end{figure}
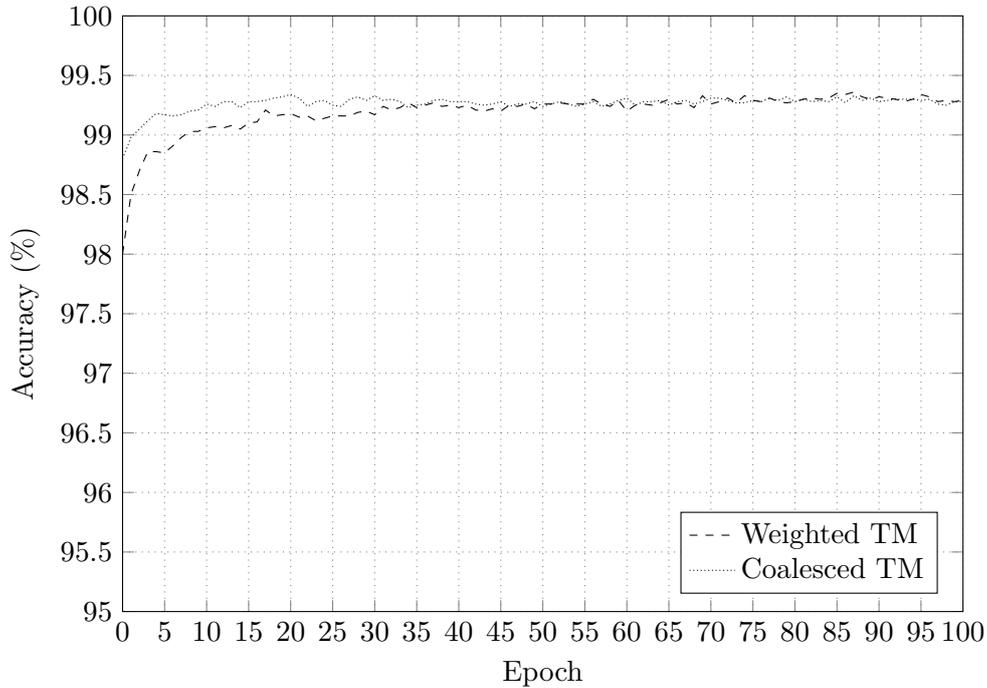

\paragraph{Learning Speed.} Figure \ref{figure:accuracy_per_epoch} plots MNIST test accuracy per epoch for \ac{CoTM} and weighted \ac{TM}. As seen, \ac{CoTM} accuracy climbs much faster, surpassing $99.3$ in epoch 19. The weighted \ac{TM} achieves this after 57 epochs, three times slower.

\paragraph{Baselines.} Table \ref{table:results} contains a comparison between the \ac{CoTM} with $64~000$ clauses and other baselines. Here, \ac{CoTM} clearly outperforms traditional machine learning techniques, like SVM, Random Forest, and Gradient Boosting. Also, it performs  competitively with widely used deep learning architectures. 
\begin{table*}[ht]
\centering
\begin{tabular}{ c|c|c|c|c } 
\hline
Model & 2D N-XOR & MNIST & K-MNIST & F-MNIST\\
\hline
\hline
4-Nearest Neighbour \cite{Clanuwat2018,xiao2017}&$61.62$&$\mathit{97.14}$&$\mathit{91.56}$&$85.40$\\
SVM \cite{Clanuwat2018}&$94.63$&$\mathit{98.57}$&$\mathit{92.82}$&$\mathit{89.7}$\\
Random Forest \cite{LiangAAAI19}&70.73&$\mathit{97.3}$&-&$\mathit{81.6}$\\
Gradient Boosting Classifier \cite{xiao2017}&87.15&$\mathit{96.9}$&-&$\mathit{88.0}$\\
Simple CNN \cite{Clanuwat2018,LiangAAAI19}&$91.05$&$\mathit{99.06}$&$\mathit{95.12}$&$\mathit{90.7}$\\
BinaryConnect \cite{courbariaux2015binaryconnect}&-&$\mathit{98.99}$&-&-\\
FPGA-accelerated BNN \cite{Lammie2019}&-&$\mathit{98.70}$&-&-\\
Logistic Circuit (binary) \cite{LiangAAAI19}&-&$\mathit{97.4}$&-&$\mathit{87.6}$\\
Logistic Circuit (real-valued) \cite{LiangAAAI19}&-&$\mathit{99.4}$&-&$\mathit{91.3}$\\
PreActResNet-18 \cite{Clanuwat2018}&-&$\mathit{99.56}$&$\mathit{97.82}$&$\mathit{92.0}$\\
ResNet18 + VGG Ensemble \cite{Clanuwat2018}&-&$\mathit{99.60}$&$\mathit{98.90}$&-\\
TM \cite{granmo2019convtsetlin}&$99.12$&$98.57$&$92.03$&$90.09$\\
\hline
\ac{CoTM} (Mean)&$99.99 \pm 0.0$&$99.31 \pm 0.0$&$96.65 \pm 0.01$&$91.83 \pm 0.01$\\
\ac{CoTM} (95 \%ile)&$100.0$&$99.34$&$96.70$&$91.93$\\
\ac{CoTM} (Peak)&$100.0$&$99.36$&$96.71$&$92.0$\\
\hline
\end{tabular}
\caption{Comparison with other machine learning techniques -- test accuracy in percent.}\label{table:results}
\end{table*}

\begin{table}[ht]
\centering
\begin{tabular}{l|c|c|c|c|c|c|c|c|c|c}
Fraction Removed & 0.0 & 0.1 & 0.2 & 0.3 & 0.4  & 0.5  & 0.6  & 0.7  & 0.8  & 0.9  \\ \hline \hline
$\Delta$ \ac{TM}         & 0.0 & 1.0 & 4.3 & 7.9 & 10.3 & 17.2 & 24.8 & 27.3 & 29.5 & 30.5 \\ \hline
$\Delta$ \ac{CoTM}        & 0.0 & 0.5 & 0.2 & 0.0 & 0.2  & 0.4  & 0.4  & 0.8  & 1.4  & 2.8
\end{tabular}
\caption{Reduction in IMDb test accuracy after removing a fraction of the training examples with positive sentiment.}\label{table:imdb}
\end{table}

\paragraph{Class Imbalance Robustness.} We here investigate whether the clause sharing of the \ac{CoTM} provides any robustness towards imbalanced training data. Table \ref{table:imdb} contains robustness results for the IMDb dataset. We evaluate robustness by removing a progressively larger fraction of the \emph{positive sentiment} training examples. The \ac{TM} test accuracy drops significantly with increasing class imbalance. However, \ac{CoTM} test accuracy is almost unaffected, indicating high robustness.

The final robustness evaluation is done with highly imbalanced CIFAR10 training data. We make the data imbalanced by ordering the classes. We then use $0.5^r$ of the training data for the class of rank $r$ in the ordering. Table \ref{table:cifar10} contains the class-wise F1-scores for \ac{TM} and \ac{CoTM}. We report the mean F1 score over epochs $90-100$, averaged over $10$ independent trials. As seen, the F1 scores of \ac{TM} are more severely affected by class imbalance than for \ac{CoTM}. 

\begin{table*}[ht]
\begin{tabular}{|l||c|c|c|c|c|c|c|c|c|c|}
Class      & airplane & automobile & bird & cat  & deer & dog  & frog & horse & ship & truck \\ \hline
\#examples & 5000      & 2500       & 1250 & 625  & 315  & 157  & 80   & 40    & 20   & 10    \\ \hline
\ac{CoTM} : F1   & 0.47     & 0.50       & 0.36 & 0.36 & 0.26 & 0.15 & 0.02 & 0.04  & 0.00 & 0.00  \\ \hline
\ac{TM} : F1    & 0.39     & 0.42       & 0.25 & 0.20 & 0.10 & 0.00 & 0.00 & 0.00  & 0.00 & 0.00
\end{tabular}
\caption{The class-wise F1 scores for a highly imbalanced version of the CIFAR10 dataset.}
\label{table:cifar10}
\end{table*}

%% file: conclusions.tex
\section{Conclusions and Further Work}\label{sec:conclusion}

In this paper, we proposed a new \ac{TM} architecture where clauses are shared among multiple outputs. We achieved this by merging multiple \acp{TM} into a single one, and then relating each clause to each output by weighting. A positive weight makes the clause vote for output $1$, while a negative weight makes it vote for output $0$. By means of a weight matrix, the clauses coalesce to produce multiple outputs.

Our empirical evaluations support the following main conclusions:
\begin{itemize}
    \item \ac{CoTM} seems to make significantly better use of few clauses, both for MNIST, Fashion-MNIST, and  Kuzushiji-MNIST. The difference is the largest for configurations with $50$ clauses per class, leading to high accuracy with frugal memory footprint and faster learning and inference.
    \item Peak accuracy is reached faster, for instance $3$ times faster on MNIST with $8~000$ clauses per class.
    \item \ac{CoTM} is highly robust towards imbalanced training data, indicating an ability to repurpose clauses from one class to another. We believe the repurposing helps classify more data sparse classes. 
\end{itemize}

In our further work, we intend to use the \ac{CoTM} to build self-supervised rule-based language models. Language models are usually based on predicting individual words or tokens from large vocabularies, requiring a large number of outputs. Further, words may have similar meanings, and we will investigate whether clause reuse can capture semantic relationships among words. We also see opportunities for using the \ac{CoTM} as an auto-encoder, supporting applications such as turning low-resolution images into high-resolution ones.

%% file: appendix.tex
\appendix

\section{Appendix}\label{appendix:a}

\subsection{Structure and Prediction}

A \ac{CoTM} can be defined as a quadruple:
\begin{equation}
\{\mathcal{X}, \mathcal{Y}, \mathcal{C}, \mathcal{W}\}.
\end{equation}
We start with defining each element of the quadruple.

\subsubsection{Input Space}

The input space of the \ac{CoTM} is denoted $\mathcal{X}$. The input space consists of vectors $\mathbf{x}$ of $o$ propositional inputs:  $\mathbf{x}=[x_1,\ldots,x_o] \in \mathcal{X}, \mathcal{X} = \{0,1\}^{o}$. 

\subsubsection{Output Space}

The output space is denoted $\mathcal{Y}$. The output space contains vectors $\mathbf{y}$ of $m$ propositional outputs: $\mathbf{y} = [y^1, y^2, \ldots, y^m] \in \mathcal{Y}, \mathcal{Y} = \{0,1\}^{m}$.

\subsubsection{Memory Matrix}

We define $\mathcal{C}$ as a space of matrices $C$ formed from $n \times 2 o$ integer entries, $C \in \mathcal{C} = \{1, 2, \ldots, 2N\}^{n \times 2o}$:
\begin{eqnarray}
C&=&
\begin{blockarray}{ccccccccc}
x_1 & x_2 &\cdots & x_o & \bar x_1 & \bar x_2 & \cdots & \bar x_o& \\
\begin{block}{[cccccccc]c}
    c_{1,1} & c_{1,2} &\cdots & c_{1,o} & c_{1,o+1} & c_{1,o+2} &\cdots & c_{1,2o} & \mathrm{clause~} 1 \\
    c_{2,1} & c_{2,2}& \cdots & c_{2,o} & c_{2,o+1} & c_{2,o+2}& \cdots & c_{2,2o} & \mathrm{clause~} 2 \\
    \vdots&\vdots&\ddots&\vdots&\vdots&\vdots&\ddots&\vdots\\
    c_{n,1} & c_{n,2}& \cdots & c_{n,o} & c_{n,o+1} & c_{n,o+2}& \cdots & c_{n,2o} & \mathrm{clause~} n\\
\end{block}
\end{blockarray}\\
&=&[c_{j,k}] \in \{1, 2, \ldots, 2N\}^{n \times 2o}.
\end{eqnarray}
This is the pattern memory of the \ac{CoTM}, with each row specifying a clause.  The $n$ matrix rows represent $n$ clauses while the $2o$ columns specifies each pattern. That is, each column refers to an input $x_k$ or its negation $\bar x_k$ (we use $\bar x_k$ as shorthand notation for $\lnot x_k$). The entries in the matrix take values from $\{1, 2, \ldots, 2N\}$, specifying the composition of the $n$ clauses as well as the confidence of the composition (how strongly the pattern is memorized).

\subsubsection{Pattern Construction}

A $C$ matrix provides $n$ patterns as follows. The first step is to map each entry in $C$ to one of two actions $a \in \{0, 1\}$, obtaining the matrix $A \in \{0,1\}^{n\times2o}$ below. The function $G(\cdot)$ performs this mapping by selecting action $0$ for entries in $C$ with values from $1$ to $N$ and by selecting action $1$ for entries with values from $N+1$ to~$2N$:
\begin{eqnarray}
g(c) &=& \begin{cases}
       0, \mathbf{if}~ 1 \le c \le N\\
       1, \mathbf{if}~ N+1 \le c \le 2N.
     \end{cases}\label{eqn:action_mapping}\\
A = G(C) &=&
\begin{bmatrix}
    g(c_{1,1}) & g(c_{1,2}) & \cdots & g(c_{1,2o})\\
    g(c_{2,1}) & g(c_{2,2}) & \cdots & g(c_{2,2o})\\
    \vdots&\vdots&\ddots&\vdots\\
    g(c_{n,1}) & g(c_{n,2}) & \cdots & g(c_{n,2o})
\end{bmatrix}\\
&=&
\begin{bmatrix}
     a_{1,1} &  a_{1,2} & \cdots &  a_{1,2o}\\
     a_{2,1} &  a_{2,2} & \cdots &  a_{2,2o}\\
    \vdots&\vdots&\ddots&\vdots\\
     a_{n,1} &  a_{n,2} & \cdots &  a_{n,2o}
\end{bmatrix}
=
[a_{j,k}] \in \{0,1\}^{n \times 2o}.
\end{eqnarray}
Values close to $N$ and $N+1$ signify low-confidence actions while values close to $1$ and $2N$ signify high-confidence actions (deeply memorized).

After the actions $A=G(C)$ have been decided, the next step is to formulate the patterns. Each pattern is formed by ANDing selected entries $x_k$ in the input vector $\mathbf{x}$, or their negations $\bar x_k$.\footnote{Such an expression is referred to as a \emph{conjunctive clause} in propositional logic, with $x_k$ and $\bar x_k$ being called \emph{literals}.} The pattern
\begin{equation}
    x_1 \land x_2 \land \bar x_3,
\end{equation}
for instance, ANDs the values $x_1$, $x_2$, and $\bar x_3$. It evaluates to $1$ if and only if $x_1=1$, $x_2=1$, and $x_3=0$.

The actions from $G(C)$ decide which inputs to AND, and whether to negate them. For $1 \le k \le o$, if action $a_{j,k}$ is $1$ then input $x_k$ takes part in the AND of pattern $j$. For $o+1 \le k \le 2o$, $\bar x_{k-o}$ takes part in the AND instead. If $a_{j,k}$ is $0$, on the other hand, the corresponding input does not play a role in the evaluation of the $j$th clause and can be ignored.

The imply ($\Rightarrow$) operator captures the above dynamics. That is, if $a_{j_k}$ is $0$, then $a_{j,k} \Rightarrow x_k$ is always $1$. Accordingly, the $x_k$ value does not impact the result. Conversely, the value of $a_{j,k} \Rightarrow x_k$ is decided solely by the $x_k$ value if $a_{j_k}$ is $1$. We thus define an $\mathit{Imply}$ operator to calculate the effect of each input on each clause:
\begin{equation}
    \mathit{Imply}[A, \mathbf{x}] = \begin{bmatrix}
    a_{1,1} \Rightarrow x_1 & \cdots & a_{1,o} \Rightarrow x_o & a_{1,o+1} \Rightarrow \bar{x}_1 & \cdots & a_{1,2o} \Rightarrow \bar x_o\\
    a_{2,1} \Rightarrow x_1 & \cdots & a_{2,o} \Rightarrow x_o & a_{2,o+1} \Rightarrow \bar x_1 & \cdots & a_{2,2o} \Rightarrow \bar x_o\\
    \vdots&\ddots&\vdots&\vdots&\ddots&\vdots\\
    a_{n,1} \Rightarrow x_1 & \cdots & a_{n,o} \Rightarrow x_o & a_{n,o+1} \Rightarrow \bar x_1 & \cdots & a_{n,2o} \Rightarrow \bar x_o
    \end{bmatrix}.
\end{equation}
Finally, we define the operator $\mathit{And}$ for evaluating the clauses:
\begin{equation}
    \mathit{And}[Z] = \begin{bmatrix}
    \bigwedge_{k = 1}^{2o} z_{1,k}\\
    \bigwedge_{k = 1}^{2o} z_{2,k}\\
    \vdots\\
    \bigwedge_{k = 1}^{2o} z_{n,k}
    \end{bmatrix}.
\end{equation}
Accordingly, the value of each clause can be calculated from the input vector $\mathbf{x}$ and the pattern memory $C$, providing a pattern value column vector $\mathbf{c}$:
\begin{equation}
    \mathbf{c} = \begin{bmatrix}
    c_1\\
    c_2\\
    \vdots\\
    c_n
    \end{bmatrix}
    =\mathit{And}(\mathit{Imply}(G(C), \mathbf{x})).
\end{equation}
As seen, we employ the $n$ patterns from $G(C)$ when processing the input vector $\mathbf{x}$.

\subsubsection{Output Prediction}

For predicting the output vector, the \ac{CoTM} uses $m \times n$ integer weights. These capture how each of the $n$ clauses relates to each of the $m$ outputs. The weights are organized as a matrix:
\begin{equation}
    W = \begin{bmatrix}
    w_{1,1} & w_{1,2} & \cdots & w_{1,n}\\
    w_{2,1} & w_{2,2} & \cdots & w_{2,n}\\
    \vdots&\vdots&\ddots&\vdots\\
    w_{m,1} & w_{m,2} & \cdots & w_{m,n}
    \end{bmatrix} = [w_{i,j}] \in \{\ldots, -2, -1, 1, 2, \ldots \}^{m \times n}.
\end{equation}
We have one weight per output $i$ per clause $j$. In the following, we will refer to row $i$ of $W$ as $\mathbf{w}_i$:
\begin{equation}
    \mathbf{w}_i = \begin{bmatrix}
    w_{i,1}&
    w_{i,2}&
    \cdots&
    w_{i,n}
    \end{bmatrix}.
\end{equation}
The weights decide the effect a clause value $c_j=1$ from $\mathbf{c}$ has on predicting the output vector. If the weight $w_{i,j}$ is positive and $c_j = 1$, the $j$th clause casts $w_{i,j}$ votes in favour of output $y_i = 1$. If the weight is negative and $c_j = 1$, clause $j$ casts $-w_{i,j}$ votes in favour of output $y_i = 0$ instead. If $c_j = 0$, on the other hand, no votes are cast from that pattern. The overall calculation can be formulated as a product between the matrix $W$ and the column vector $\mathbf{c}$, obtaining a column vector $\mathbf{v}$ of votes:
\begin{equation}\label{eqn:vote_sum}
    \mathbf{v} = W \mathbf{c}.
\end{equation}
Finally, the output vector prediction is decided by a majority vote using a unit step operator~$U$:
\begin{equation}
    U(\mathbf{v}) = \begin{bmatrix}
    v_1 \ge 0\\
    v_2 \ge 0\\
    \vdots\\
    v_m \ge 0
    \end{bmatrix} = [u_i] \in \{0, 1\}^{m \times 1}.
\end{equation}
Combining the different steps, we predict the output $\mathbf{y}$ from input $\mathbf{x}$ as follows:
\begin{equation}
    \hat{\mathbf{y}} = U(W \cdot \mathit{And}(\mathit{Imply}(G(C), \mathbf{x})))^T.\label{eqn:prediction}
\end{equation}

\subsection{Learning Patterns and Weights}

The memory matrix $C$ and the weight matrix $W$ are learnt online from a set $S$ of training examples $(\mathbf{x}, \mathbf{y}) \in S$. The goal is to maximize prediction accuracy $P(\hat{y}_i = y_i)$ per entry in $\mathbf{y}$. We here consider the procedure for updating $C$ and $W$ from a single training example $(\mathbf{x}, \mathbf{y})$. To learn from a stream of training examples, this procedure is repeated for every example. We first extend the \ac{TM} clause coordination mechanism to stimulate production of diverse clauses for multiple outputs. This mechanism activates clause rows in $C$ for updating based on the vote sums $\mathbf{v}$ and the example outputs $\mathbf{y}$. We then go through the details of how we update each activated clause in $C$ based on $(\mathbf{x}, \mathbf{y})$. Finally, we consider the updating of the corresponding weights in~$W$, learning the relationship between each clause in $C$ and each output in $\mathbf{y}$.

\subsubsection{Resource Allocation}

To ensure that the $n$ clause rows of $C$ are diverse and robust, the \ac{CoTM} employs a voting margin $t$ for each entry $y_i$ in $\mathbf{y}$. The voting margin is a user-configurable constant that puts a max on how many clauses from $C$ coalesce to predict a particular output $y_i$ when facing a training example $(\mathbf{x}, \mathbf{y})$. While the original \ac{TM} considers a single output $y$, we here present a scheme for handling a vector $\mathbf{y}$ of outputs.

For \ac{CoTM}, when an entry $y_i$ in the training output $\mathbf{y}$ is $0$, the summation margin is $-t$. Conversely, when the entry is $1$ the summation margin is $t$. The resulting $n$ summation margins for $\mathbf{y}$ are organized in the column vector $\mathbf{q}$:
\begin{equation}
\mathbf{q} = \mathbf{y} t - \bar{\mathbf{y}} t =
\begin{bmatrix}
y_1 t - \bar y_1 t\\
y_2 t - \bar y_2 t\\
\vdots\\
y_m t - \bar y_m t
\end{bmatrix}
= [q_i] \in \{-t, t\}^{m \times 1}.
\end{equation}

After the summation margins $\mathbf{q}$ for the training example $(\mathbf{x}, \mathbf{y})$ have been set, the next step is to decide which clause rows in $C$ to update. This is done by comparing each vote sum $v_i$ from $\mathbf{v}$ (Eqn. \ref{eqn:vote_sum}) with the corresponding margin $y_i t - \bar y_i t$ in $\mathbf{q}$. By clipping the vote sum to fall between $-t$ and $t$, we calculate an update probability per output $y_i$ in $\mathbf{y}$: $\mathit{abs}\left(\frac{q_i - \mathit{clip}(v_i, -t, t)}{2t}\right)$. The update probability is $0.0$ when the summation margin $t$ is met or surpassed, and gradually increases to $1.0$ towards the maximum distance $2t$. The update probabilities are organized in the column vector $\mathbf{d}$:
\begin{equation}
\mathbf{d} = \mathit{abs}(\mathbf{q} - \mathit{clip}(\mathbf{v}, -t, t)) \cdot \frac{1}{2t} = \mathit{abs} \left( \begin{bmatrix}
\frac{q_1 - \mathit{clip}(v_1, -t, t)}{2t} \\
\frac{q_2 - \mathit{clip}(v_2, -t, t)}{2t}\\
\vdots\\
\frac{q_m - \mathit{clip}(v_m, -t, t)}{2t}
\end{bmatrix}\right)
= [d_i] \in \{x \in R | 0 \le x \le 1\}^{m \times 1}.
\end{equation}
From the update probabilities in $\mathbf{d}$, we next randomly select which clause rows in $C$ to update. In brief, many clauses are updated when the vote sum $v_i$ is far from its margin $y_i t - \bar y_i t$, and comes to a complete standstill when the margin is met or surpassed. Accordingly, we attain a natural balancing of pattern representation resources.

\subsubsection{Type I Feedback -- Suppressing False Negative Output}

There are two kinds of updates, Type I and Type II. We consider the nature of these next. \emph{Type I Feedback} reinforces the output entries $y_i$ from the training example  $(\mathbf{x}, \mathbf{y})$, considering one output entry $y_i$ at a time. Each output entry $y_i$ is either $0$ or $1$. As seen from Eqn. \ref{eqn:prediction}, output $y_i=1$ can be reinforced by identifying the clause rows $j$ in $C$ that has a positive weight $w_{i,j} \ge 0$, and then making those evaluate to $1$. Output $y_i = 0$, on the other hand, can be reinforced by making clauses with negative weight $w_{i,j} < 0$ evaluate to $1$. In other words, Type~I Feedback is given when $y_i \odot (w_{i,j} \ge 0)$, where $\odot$ is the XNOR-operator.

The selection of clauses to update is done by sampling a random value $0$ or $1$ per output-clause pair $(i,j)$, with $i$ referring to an entry in $\mathbf{y}$ and $j$ referring to a row in $C$. The probability of updating from a pair $(i,j)$ is decided by the update probability $d_i$ in $\mathbf{d}$. We draw $m\times n$ random values $\pi_{i,j}$ uniformly in the range $[0,1]$ to decide the updating. The updating performed per pair is the same as for the original \ac{TM}, except that the \ac{CoTM} combines multiple outputs~$y_i$.

We organize the selection criterion $y_i \odot (w_{i,j} \ge 0)$ in one $m \times n$ matrix and the random sampling in another. The two matrices are combined by element-wise multiplication using the Hadamard product $\circ$:  \begin{eqnarray}
R^I&=& 
\begin{bmatrix}
y_1 \odot (w_{1,1} \ge 0)&y_1 \odot (w_{1,2} \ge 0)&\cdots&y_1 \odot (w_{1,n} \ge 0)\\
y_2 \odot (w_{2,1} \ge 0)&y_1 \odot (w_{2,2} \ge 0)&\cdots&y_2 \odot (w_{2,n} \ge 0)\\
\vdots&\vdots&\ddots&\vdots\\
y_m \odot (w_{m,1} \ge 0)&y_1 \odot (w_{m,2} \ge 0)&\cdots&y_m \odot (w_{m,n} \ge 0)
\end{bmatrix}
\circ\\
&&\begin{bmatrix}
\pi_{1,1} \le d_1 &\pi_{1,2} \le d_1&\cdots&\pi_{1,n} \le d_1\\
\pi_{2,1} \le d_2&\pi_{2,2} \le d_2&\cdots&\pi_{2,n} \le d_2\\
\vdots&\vdots&\ddots&\vdots\\
\pi_{m,1} \le d_m&\pi_{m,2} \le d_m&\cdots&\pi_{m,n} \le d_m
\end{bmatrix}\\
&=& [r^I_{i,j}] \in \{0,1\}^{m \times n}.
\end{eqnarray}
In other words, matrix $R^I$ picks out which rows in $C$ to update, per output $y_i$. Like for the original \ac{TM}, the actual updates subdivides into Type Ia and Type Ib feedback, which we consider next.

\paragraph{Type Ia Feedback -- Refining Clauses.} Type Ia Feedback is given to clause rows $j$ in $C$ that evaluate to $1$. Reinforcement is given per variable $x_k$ or its negation $\bar x_k$, by updating matrix $C$. In brief, action $a=1$ is reinforced by increasing $c_{j,k}$ whenever $c_j =1~\mathbf{and}~x_k=1$ for $1 \ge k \ge o$ and $c_j =1~\mathbf{and}~ \bar x_{k-o}=1$ for $o+1 \ge k \ge 2o$. This reinforcement is strong and makes the clause remember and refine the pattern it recognizes in $\mathbf{x}$.\footnote{Note that we here apply boosting of true positives. The standard setup is to apply the update with probability $\frac{s-1}{s}$.} Which entries $c_{j,k}$ in $C$ to increase in this manner are organized in the matrix $F^{Ia}_i$, per output $y_i$:
\begin{eqnarray}
F^{Ia}_i &=&
\begin{bmatrix}
\mathbf{c}&\cdots&\mathbf{c}\\
\end{bmatrix}
\circ
\begin{bmatrix}
x_1 &\cdots&x_o&\bar x_1 &\cdots&\bar x_o\\
x_1 &\cdots&x_o&\bar x_1 &\cdots&\bar x_o\\
\vdots&\ddots&\vdots&\vdots&\ddots&\vdots\\
x_1 &\cdots&x_o&\bar x_1 &\cdots&\bar x_o\\
\end{bmatrix}\\
&=&
\begin{bmatrix}
c_1&c_1&\cdots&c_1\\
c_2&c_2&\cdots&c_2\\
\vdots&\vdots&\ddots&\vdots\\
c_n&c_n&\cdots&c_n
\end{bmatrix}
\circ
\begin{bmatrix}
x_1 &\cdots&x_o&\bar x_1 &\cdots&\bar x_o \\
x_1 &\cdots&x_o&\bar x_1 &\cdots&\bar x_o \\
\vdots&\ddots&\vdots&\vdots&\ddots&\vdots\\
x_1 &\cdots&x_o&\bar x_1 &\cdots&\bar x_o \\
\end{bmatrix}.
\end{eqnarray}

\paragraph{Type Ib Feedback -- Coarsening Clauses.} Type Ib Feedback reinforces actions $a_{j,k}=0$. That is, $a_{j,k}=0$ in matrix $A$  is reinforced by reducing $c_{j,k}$ in matrix $C$. The reinforcement happens randomly with probability $\frac{1}{s}$ whenever $c_j =0~\mathbf{or}~x_k=0$ for $1 \ge k \ge o$ and whenever $c_j =0~\mathbf{or}~\bar x_{k-o}=0$ for $o+1 \ge k \ge 2o$. This reinforcement is weak (triggered with low probability) and coarsens infrequent patterns, making them frequent. The user-configurable parameter $s$ controls pattern frequency, i.e., a higher $s$ produces less frequent patterns.  Which entries $c_{j,k}$ in $C$ to reduce in this manner are organized in the matrix $F^{Ib}_i$, per output entry $y_i$.

We first randomly identify which entries in $C$ are candidates for updating.  We draw $n \times 2o$ random values $\pi_{i,j}$ uniformly in the range $[0,1]$ to decide this updating, organized in the $n \times 2o$ matrix $B^i$:
\begin{equation}
B^i = \begin{bmatrix}
\pi_{1,1} \le \frac{1}{s}&\pi_{1,2} \le \frac{1}{s}& \cdots & \pi_{1,2o} \le\frac{1}{s}\\
\pi_{2,1} \le \frac{1}{s}&\pi_{2,2} \le \frac{1}{s}& \cdots & \pi_{2,2o}\le \frac{1}{s}\\
\vdots&\vdots& \ddots & \vdots\\
\pi_{n,1} \le \frac{1}{s}&\pi_{n,2} \le \frac{1}{s}& \cdots & \pi_{n,2o} \le \frac{1}{s}
\end{bmatrix}
= [b_{j,k}^i] \in \{0,1\}^{n \times 2o}.
\end{equation}
In this manner, which entries in $C$ to decrease is organized in the $m \times 2o$ matrix $F^{Ib}_i$ for the particular output $y_i$:
\begin{eqnarray}
F^{Ib}_i&=&
B_i \circ 
\begin{bmatrix}
\bar x_1 \lor \bar c_1 &\cdots&\bar x_o \lor \bar c_1&x_1 \lor \bar c_1 &\cdots&x_o \lor \bar c_1\\
\bar x_1 \lor \bar c_2 &\cdots&\bar x_o \lor \bar c_2&x_1 \lor \bar c_2 &\cdots&x_o \lor \bar c_2\\
\vdots&\ddots&\vdots&\vdots&\ddots&\vdots\\
\bar x_1 \lor \bar c_n &\cdots&\bar x_o \lor \bar c_n&x_1 \lor \bar c_n &\cdots&x_o \lor \bar c_n
\end{bmatrix}.
\end{eqnarray}

\subsubsection{Type II Feedback -- Increasing Discrimination Power}

Type II Feedback is given to clause rows $j$ in $C$ that outputs false positives. That is, when $y_i=1$ the afflicted clauses are the ones with negative weight $w_{i,j} < 0$, considering each output-clause pair (j,i). Conversely, when $y_i=0$ clauses with positive weight $w_{i,j} \ge 0$ are targeted. More concisely stated,  Type II Feedback is given when $y_i \oplus (w_{i,j} \ge 0)$, where $\oplus$ is the XOR-operator.

In addition to this selection criteria, rows $j$ in $C$ are selected randomly for updating per pair $(i,j)$ decided by the update probability $d_i$ in $\mathbf{d}$. Uniquely for Type II Feedback, $\mathbf{d}$ is further multiplied by a scalar $e$, which usually is set to $1.0$. However, for one-hot-encoded multi-class problems, $e$ is set to $1.0/(m-1)$. The two selection criteria are combined through element-wise multiplication using the Hadamard product, and organized in the $m \times n$ matrix $R^{II}$:
\begin{eqnarray}
R^{II}&=& 
\begin{bmatrix}
y_1 \oplus (w_{1,1} \ge 0)&y_1 \oplus (w_{1,2} \ge 0)&\cdots&y_1 \oplus (w_{1,n} \ge 0)\\
y_2 \oplus (w_{2,1} \ge 0)&y_1 \oplus (w_{2,2} \ge 0)&\cdots&y_2 \oplus (w_{2,n} \ge 0)\\
\vdots&\vdots&\ddots&\vdots\\
y_m \oplus (w_{m,1} \ge 0)&y_1 \oplus (w_{m,2} \ge 0)&\cdots&y_m \oplus (w_{m,n} \ge 0)
\end{bmatrix}
\circ\\
&&\begin{bmatrix}
\pi_{1,1} \le d_1 e &\pi_{1,2} \le d_1 e&\cdots&\pi_{1,n} \le d_1 e\\
\pi_{2,1} \le d_2 e&\pi_{2,2}\le d_2 e&\cdots&\pi_{2,n} \le d_2 e\\
\vdots&\vdots&\ddots&\vdots\\
\pi_{m,1} \le d_m e&\pi_{m,2} \le d_m e&\cdots&\pi_{m,n} \le d_m e
\end{bmatrix}\\
&=& [r^{II}_{i,j}] \in \{0,1\}^{m \times n}.
\end{eqnarray}

Like Type Ia Feedback,  we only update clause rows $j$ in $C$ that evaluates to $1$, i.e., $c_j = 1$. Finally, only \emph{Exclude} actions for literals of value $0$ are given feedback, reinforcing \emph{Include}. That is, state $c_{j,k}$ increases when $\bar x_k \bar a_{j,k} = 1$, for $1 \ge k \ge o$. For $o+1 \ge k \ge 2o$, state $c_{j,k}$ increases when $x_{k-o} \bar a_{j,k} = 1$. These criteria are ANDed in the $F^{II}_i$ matrix through element-wise multiplication:
\begin{eqnarray}
F^{II}_i &=&
\begin{bmatrix}
\mathbf{c}&\cdots&\mathbf{c}\\
\end{bmatrix}
\circ
\begin{bmatrix}
\bar x_1 \bar a_{1,1} &\cdots&\bar x_o \bar a_{1,o} &x_1 \bar a_{1,o+1}&\cdots&x_o \bar a_{1,2o}\\
\bar x_1 \bar a_{2,1} &\cdots&\bar x_o \bar a_{2,o} &x_1 \bar a_{2,o+1}&\cdots&x_o \bar a_{2,2o}\\
\vdots&\ddots&\vdots&\vdots&\ddots&\vdots\\
\bar x_1 \bar a_{n,1}&\cdots&\bar x_o \bar a_{n,o}&x_1 \bar a_{n,o+1}&\cdots&x_o \bar a_{n,2o}
\end{bmatrix}\\
&=&
\begin{bmatrix}
c_1&c_1&\cdots&c_1\\
c_2&c_2&\cdots&c_2\\
\vdots&\vdots&\ddots&\vdots\\
c_n&c_n&\cdots&c_n
\end{bmatrix}
\circ
\begin{bmatrix}
\bar x_1 \bar a_{1,1} &\cdots&\bar x_o \bar a_{1,o} &x_1 \bar a_{1,o+1}&\cdots&x_o \bar a_{1,2o}\\
\bar x_1 \bar a_{2,1} &\cdots&\bar x_o \bar a_{2,o} &x_1 \bar a_{2,o+1}&\cdots&x_o \bar a_{2,2o}\\
\vdots&\ddots&\vdots&\vdots&\ddots&\vdots\\
\bar x_1 \bar a_{n,1}&\cdots&\bar x_o \bar a_{n,o}&x_1 \bar a_{n,o+1}&\cdots&x_o \bar a_{n,2o}
\end{bmatrix}
\end{eqnarray}

\subsubsection{Updating the Memory Matrix}

The memory matrix $C$ is updated based the training example $(\mathbf{x}, \mathbf{y})$ using the Type I and Type II feedback matrices $F_i^{Ia}$, $F_i^{Ia}$, and $F_i^{II}$. This updating is done per output $y_i$ in $\mathbf{y}$, simply by summation:
\begin{equation}
    C^*_{T+1} = C_T + \sum_{i=1}^m \left(Q^{II}_i \circ F^{II}_i + Q^I_i \circ F^{Ia}_i - Q^I_i \circ F^{Ib}_i\right).
\end{equation}
Above, $Q^I_i$ maps $R^{I}_i$ down to the literals:
\begin{equation}
Q^I_i = \begin{bmatrix}
\mathbf{r}_i^{I}\\
\vdots\\
\mathbf{r}_i^{I}\\
\end{bmatrix}^T
=
\begin{bmatrix}
r_{i,1}^I&r_{i,1}^I&\cdots&r_{i,1}^I\\
r_{i,2}^I&r_{i,2}^I&\cdots&r_{i,2}^I\\
\vdots&\vdots&\ddots&\vdots\\
r_{i,n}^I&r_{i,n}^I&\cdots&r_{i,n}^I
\end{bmatrix}.
\end{equation}
Correspondingly, $Q^{II}_i$ deals with $R^{II}_i$:
\begin{equation}
Q^{II}_i = \begin{bmatrix}
\mathbf{r}_i^{II}\\
\vdots\\
\mathbf{r}_i^{II}\\
\end{bmatrix}^T
=
\begin{bmatrix}
r_{i,1}^{II}&r_{i,1}^{II}&\cdots&r_{i,1}^{II}\\
r_{i,2}^{II}&r_{i,2}^{II}&\cdots&r_{i,2}^{II}\\
\vdots&\vdots&\ddots&\vdots\\
r_{i,n}^{II}&r_{i,n}^{II}&\cdots&r_{i,n}^{II}
\end{bmatrix}.
\end{equation}
The final step is to clip the state values in $C$ to make sure that they stay within $1$ and $2N$:
\begin{equation}
    C_{t+1} = \mathit{clip}\left(C^*_{t+1}, 1, 2N\right).
\end{equation}

\subsubsection{Learning Weights}

The perhaps most novel part of \ac{CoTM} is the fact that the clauses are not initially related to any particular output $y_i$ in $\mathbf{y}$. By learning how to relate each clause to each output, the \ac{CoTM} opens up for clause sharing. That is, the weight matrix $W$ is initialized randomly with values from $\{-1, 1\}$, thus randomly assigning each clause to a truth value for $y_i$. The original \ac{TM}, on the other hand, operates with a fixed weight matrix $W$ where each clause is preassigned to a particular output. That is, the clause rows are first partitioned into $m$ partitions of equal size. Half of the clause rows inside a partition is then assigned the weight $-1$ and the other half is assigned the weight $+1$. For other outputs the weight is set to $0$, thus, blocking sharing of clauses between the outputs.

The updating of weights is quite straightforward. Type Ia Feedback changes the weight in one-step increments away from $0$, while Type II Feedback changes the weight in one-step decrements towards $0$. Type Ib Feedback leaves the weights unchanged. This updating of the weight matrix $W$ from  $W_t$ to $W_{t+1}$ is captured by the following equation:
\begin{equation}
W_{t+1} = W_t + (R^I + R^{II}) \circ
\left(\begin{bmatrix}
c_1 y_1&c_2 y_1&\cdots&c_n y_1 \\
c_1 y_2&c_2 y_2&\cdots&c_n y_2\\
\vdots&\vdots&\ddots&\vdots\\
c_1 y_m&c_2 y_m&\cdots&c_n y_m\\
\end{bmatrix}-
\begin{bmatrix}
c_1 \bar y_1&c_2 \bar y_1&\cdots&c_n \bar y_1\\
c_1 \bar y_2&c_2 \bar y_2&\cdots&c_n \bar y_2\\
\vdots&\vdots&\ddots&\vdots\\
c_1 \bar y_m& c_2 \bar y_m &\cdots&c_n \bar y_m
\end{bmatrix}\right).
\end{equation}
Note that the $1$-entries of $R^I$ and $R^{II}$ are non-overlapping, singling out which clause rows $j$ are eligible for Type Ia and Type II Feedback. The value of $y_i$ controls the direction of the weight update.